\documentclass[journal]{IEEEtran}
\usepackage{array}

\usepackage{multirow}
\usepackage{color}
\usepackage[pdftex]{graphicx}
\DeclareGraphicsExtensions{.pdf,.jpeg,.png}
\usepackage{epstopdf}

\ifCLASSINFOpdf
   \usepackage[pdftex]{graphicx}
   \usepackage{subfigure}
\else
\fi
\usepackage{amsmath}
\usepackage{algorithm}
\usepackage{algorithmicx}
\usepackage{algpseudocode}

\usepackage[numbers]{natbib}

\begin{document}
%
\title{PaDNet: Pan-Density Crowd Counting}
%
%
%

\author{Yukun Tian,
        Yiming Lei,
        Junping~Zhang,~\IEEEmembership{Member,~IEEE,}
        and James Z. Wang
\thanks{Y. Tian, Y. Lei and J. Zhang  are with the Shanghai Key Laboratory
of Intelligent Information Processing, School of Computer Science,
Fudan University, Shanghai 200433, China (e-mails: \{yktian17, ymlei17, jpzhang\}@fudan.edu.cn).}
\thanks{J. Z. Wang is with the College of Information Sciences and Technology,
The Pennsylvania State University, University Park, PA 16802, USA (e-mail:
jwang@ist.psu.edu).}

\thanks{Manuscript received February 27, 2019; revised September 20, 2019; accepted October 22, 2019{\sl (Corresponding author: Junping~Zhang.)}}
}

\maketitle


\begin{abstract}
Crowd counting is a highly challenging problem in computer vision and machine learning. Most previous methods have focused on consistent density crowds, {\it i.e.}, either a sparse or a dense crowd, meaning they performed well in global estimation while neglecting local accuracy. To make crowd counting more useful in the real world, we propose a new perspective, named pan-density crowd counting, which aims to count people in varying density crowds. Specifically, we propose the Pan-Density Network (PaDNet) which is composed of the following critical components. First, the Density-Aware Network (DAN) contains multiple subnetworks pretrained on scenarios with different densities. This module is capable of capturing pan-density information. Second, the Feature Enhancement Layer (FEL) effectively captures the global and local contextual features and generates a weight for each density-specific feature. Third, the Feature Fusion Network (FFN) embeds spatial context and fuses these density-specific features. Further, the metrics Patch MAE (PMAE) and Patch RMSE (PRMSE) are proposed to better evaluate the performance on the global and local estimations. Extensive experiments on four crowd counting benchmark datasets, the ShanghaiTech, the UCF\_CC\_50, the UCSD, and the UCF-QNRF, indicate that PaDNet achieves state-of-the-art recognition performance and high robustness in pan-density crowd counting.
\end{abstract}


%
\IEEEpeerreviewmaketitle

\section{\textbf{Introduction}}

\IEEEPARstart{C}{rowd} counting has broad applications in public safety, emergency evacuation, smart city planning, and news reporting~\cite{ford2017trump}. However, due to perspective distortions, severe occlusions, high-variant densities, and other problems, crowd counting has been a persistent challenge in computer-vision and machine-learning domains. Existing work has largely focused on consistent density crowds, {\it i.e.}, either a sparse or a dense crowd. Yet in the real world, an image of a crowd may have areas of inconsistent densities due to camera perspective as well as naturally varying distribution of people in the crowd. Accurately counting people in crowds, thus, entails consideration of all density variations. To emphasize this density-inclusive focus, we name our approach {\it pan-density crowd counting}.

\begin{figure}[t]
 \centering
 \includegraphics[width=\linewidth, clip=true, trim=200 100 250 90]{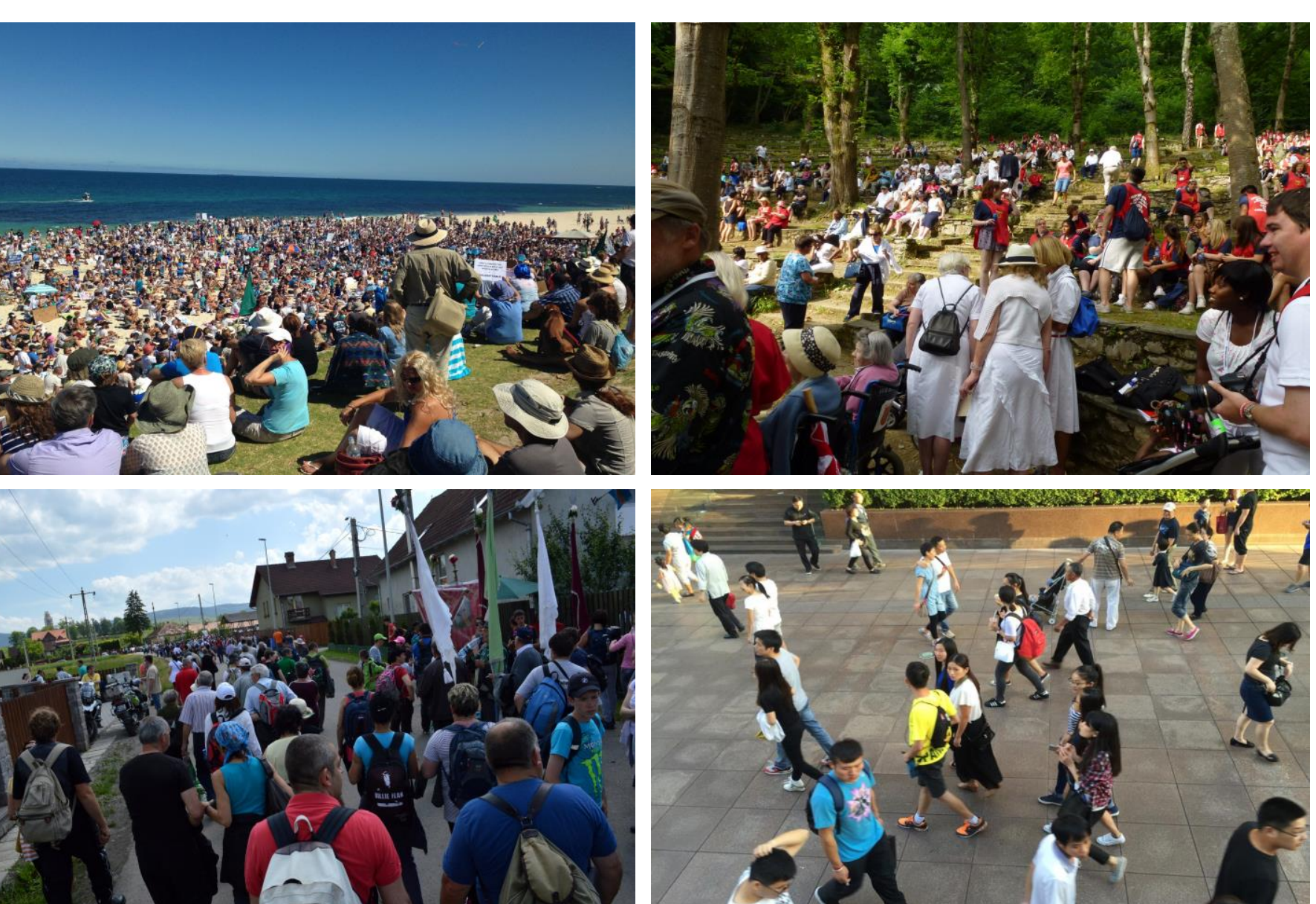}
 \caption{Example crowd images. The last image is from the ShanghaiTech Part\_B dataset~\cite{Zhang2016Single} and the other images are from the UCF-QNRF dataset~\cite{idrees2018composition}. These images show crowds of diverse densities and distributions. Furthermore, the densities of local regions could be inconsistent  in the same scene.}
 \label{fig1}
\end{figure}

The needs for pan-density crowd counting is evident in the crowd image examples shown in Figure~\ref{fig1}. The images convey two properties: (i) different crowds have varying densities and distributions, and more importantly, (ii) the densities of local regions can be inconsistent even in the same scene. Because prior methods focus on the global evaluation in varying-density scenes, they cannot sufficiently capture pan-density information to achieve good performance in terms of local accuracy. Their recognition accuracy is therefore limited in dealing with pan-density crowd counting.

Some earlier methods count sparse pedestrians by using a sliding window detector~\cite{1541243,5995698}. Regression-based approaches~\cite{chan2009bayesian,ryan2009crowd} utilize hand-crafted features extracted from local image patches to count sparse crowds.
Due to severe occlusions, these methods have  limited performance in dense crowd counting.
Inspired by the success of convolutional neural network\;(CNN)~\cite{8576646,8576656,7839189,chao2019gaitset}, researchers employed CNN-based methods to predict a density map which includes important spatial distribution information for dense crowd counting.
\begin{figure*}[ht]
 \centering
  \subfigure[Original]{\includegraphics[width=0.325\linewidth]{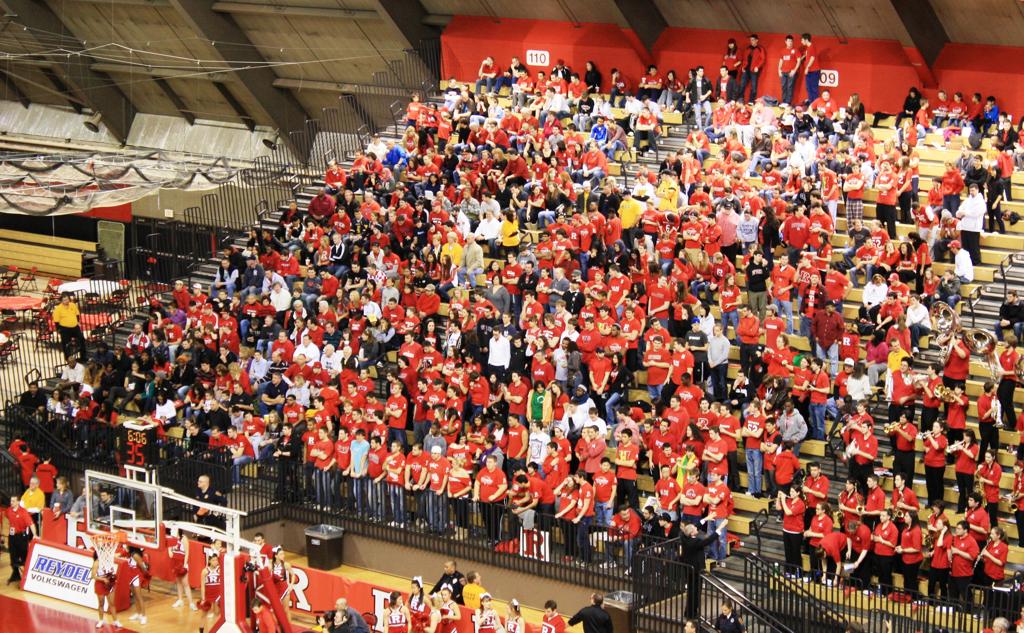}}
  \subfigure[GT Count: 553]{\includegraphics[width=0.325\linewidth, height=0.20\linewidth]{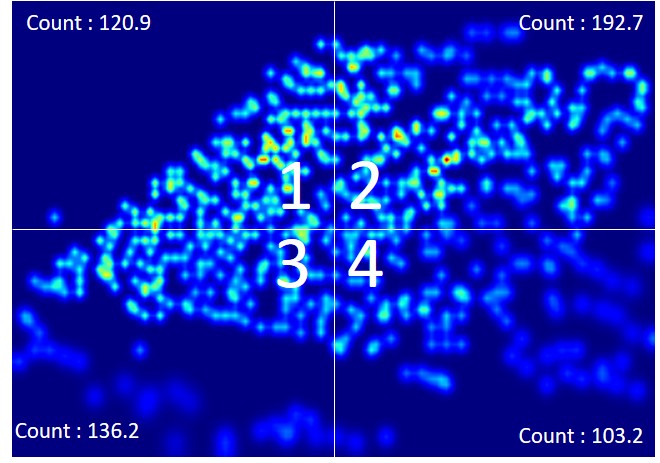}}
  \subfigure[MCNN Count: 557.2]{\includegraphics[width=0.325\linewidth, height=0.20\linewidth]{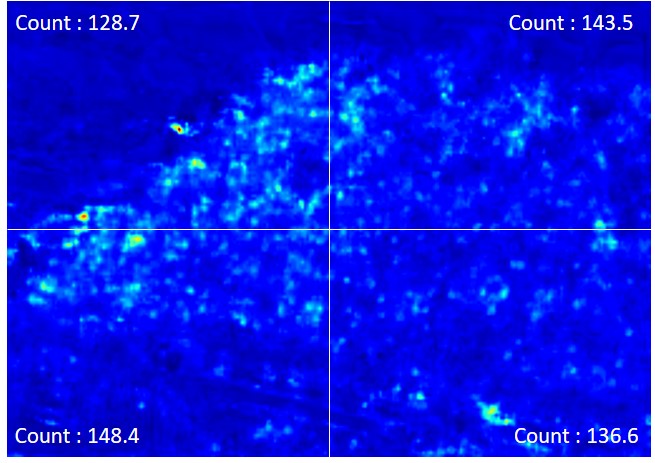}}
 \caption{The original image is from the ShanghaiTech Part\_A dataset~\cite{Zhang2016Single}. The ground truth is shown in (b). The density map generated by MCNN in shown in (c). The global estimation of MCNN is close to the ground truth, but the local estimation is biased.
Accurate global estimation, however, is because the underestimation of region 2 offsets the overestimation of regions 3 and 4.}
 \label{fig2}
\end{figure*}
Single-column CNNs~\cite{wang2015deep,Fu2015Fast} adopt multiple convolutional layers to extract features, and these features are fed into a fully connected layer to count people in dense scenes. While these single-column-based methods are suitable for single-density crowd counting, they cannot fully capture pan-density information.

In order to count crowds with varying densities, multi-column-based network methods have been developed~\cite{Zhang2016Single,8296324,sam2017switching,sindagi2017generating}. These methods contain several columns of CNNs that have different-sized filters to capture multi-scale information. For instance, filters with larger receptive fields are more useful for modeling the density maps corresponding to larger head regions. However, these multi-scale-based methods have relatively low efficiency because they cannot accurately recognize a specific density crowd or reasonably utilize the features learned by networks across columns.

Li \textit{et al.}~\cite{li2018csrnet} shows that because these methods cannot accurately learn different features for each column, they result in some ineffective and redundant branches. To address this issue, Sam \textit{et al.}~\cite{sam2017switching} proposed a Switch-CNN through training the switch classifier to select the optimal regressor for one input patch. Yet, Switch-CNN is limited because it chooses one of the results of different subnetworks rather than fusing them. High-variant density exists not only at the whole image level, but also at the image patch level. Each subnetwork of Switch-CNN is trained on a specific density subdataset and thus cannot utilize the whole dataset. Therefore, the single subnetwork has  limited recognition performance and cannot overcome the feature covariate shift problem.

 Most of these multi-column-based models fuse the feature maps generated by different columns  via a $1\times1$ convolutional layer. As a result, the operation suffers from multi-scale model competition. A more reasonable way of fusing feature maps is to assign different weights for the subnetworks. Sindagi \textit{et al.}~\cite{sindagi2017generating} proposed a Contextual Pyramid
CNN\;(CP-CNN) to incorporate contextual information for achieving low counting error and high-quality density maps.
This approach, however, has a high computation complexity in predicting the global and local contexts. Further, predicting the local context is a difficult task, and once the prediction is biased, overall performance is severely limited.

In addition, although they are accurate in estimating the global count in the scene, the fatal flaw of most crowd-counting methods is the bias of their local estimations. An example in Figure~\ref{fig2} shows that the estimation\;(=557.2) generated by Multi-Column CNN\;(MCNN)~\cite{Zhang2016Single}, which is a representative crowd-counting algorithm, is close to the ground truth of 553. However, the local estimations are quite inaccurate. For example, the estimation of MCNN in the region 2 is 143.5, while the ground truth is 192.7. There also exists relatively large biases in the regions 3 and 4. By observing the biased local estimation, it appears that the high accuracy of global estimation stems from a fact that the underestimation of region 2 offsets the overestimation of regions 3 and 4. It can also be seen that two general evaluation metrics of crowd counting, MAE and RMSE, prefer estimating global accuracy and robustness over estimating local ones.

In order to tackle the aforementioned problems, we propose a novel model, called PaDNet, for pan-density crowd counting. PaDNet is composed of three critical components. First, the Density-Aware Network (DAN) contains multiple subnetworks pretrained on scenarios with different densities. This module is capable of capturing the pan-density information. Second, the Feature Enhancement Layer (FEL) effectively captures the global and local contextual features and generates a weight for each density-specific feature. Third, the Feature Fusion Network (FFN) embeds spatial context and fuses these density-specific features instead of choosing one.

Our {\bf main contributions} are summarized below.
\begin{itemize}
\item We propose a novel end-to-end architecture named PaDNet for pan-density crowd counting. Further, we explore the impact of density-level division on estimation performance. Through extensive experiments on four benchmark crowd datasets, PaDNet obtains the best performance and high robustness in pan-density crowd counting compared with state-of-the-art algorithms.
\item In order to evaluate both local accuracy and robustness, we propose two new evaluation measures, {\it i.e.}, Patch MAE\;(PMAE) and Patch RMSE\;(PRMSE). They consider both global accuracy and robustness as well as local ones.
\end{itemize}

The remainder of the paper is organized as follows. Section~\ref{sec:rel} introduces related works in crowd counting. In Section~\ref{sec:met}, we present the details of our method and then we present and analyze the experimental results in Section~\ref{sec:exp}. We offer final thoughts in Section~\ref{sec:con}.

\begin{figure*}[htbp]
 \centering
  \includegraphics[width=\linewidth]{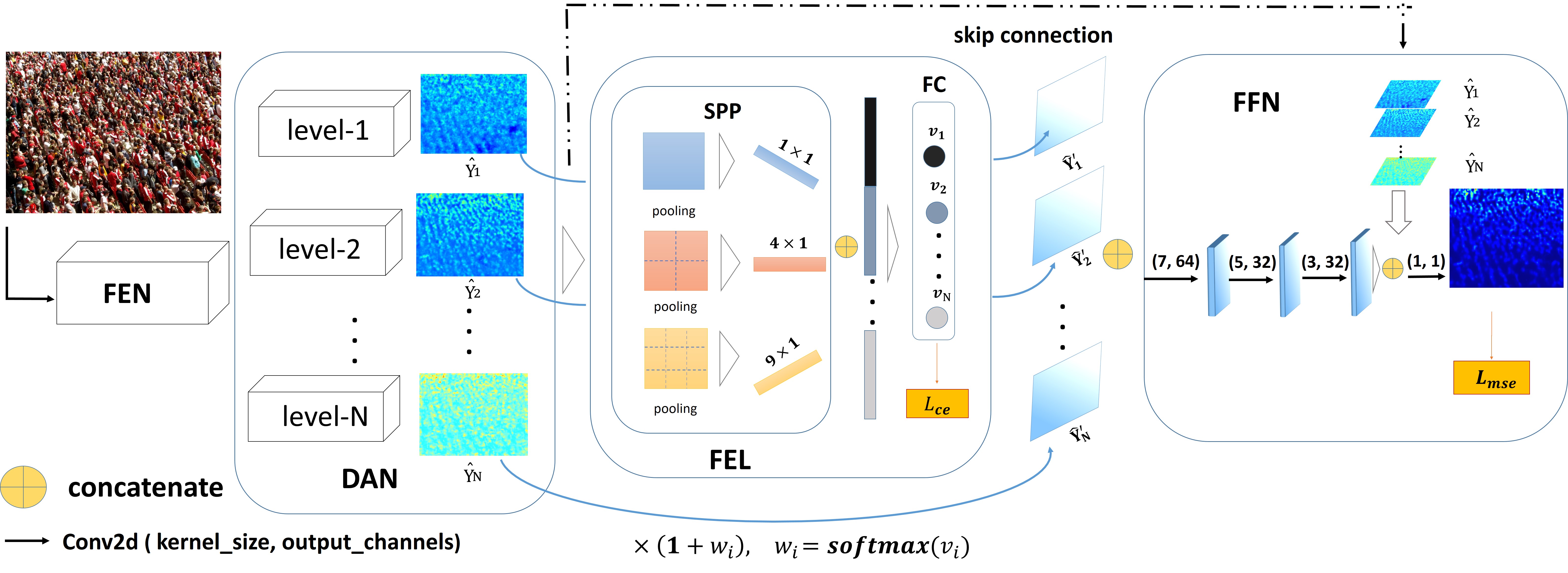}
 \caption{The PaDNet consists of Feature Extraction Network\;(FEN), Density-Aware Network\;(DAN), Feature Enhancement Layer\;(FEL), and Feature Fusion Network\;(FFN). FEN extracts low-level feature of images. DAN employs multiple subnetworks to recognize different density levels in crowds and to generate the feature map ($\hat{Y}_i$). FEL captures the global and local features and learns an enhancement rate to boost the feature map $\hat{Y}_i$ and generates $\hat{Y}^{\prime}_i$. FFN fuses $\hat{Y}^{\prime}_i$ and generates the final density map for counting.}
 \label{fig3}
\end{figure*}

%
%
%
%

\section{\textbf{Related Works}}\label{sec:rel}
Existing crowd counting algorithms can be roughly categorized into detection-based methods, regression-based methods, and CNN-based methods. Below we give a brief survey of these three categories.
\subsection{\textbf{Detection-based Methods}}
Detection-based methods of crowd counting utilize a moving-window detector to identify pedestrians and count the number of people in an image~\cite{Pedestrian}. Some researchers have proposed extracting common features from appearance-based crowd images in order to count crowds~\cite{dalal2005histograms,leibe2005pedestrian,tuzel2008pedestrian}, but these approaches have obtained limited recognition performance improvement when dealing with dense crowd counting. To overcome this issue, researchers used part-based methods to detect the specific body parts such as the head or the shoulder to count pedestrians~\cite{felzenszwalb2010object,wu2007detection}. However, these detection-based methods are only suitable for counting sparse crowds because they are affected by severe occlusions.


\subsection{\textbf{Regression-based Methods}}
To address the problem of occlusion, regression-based methods have been introduced for crowd counting. The main idea of regression-based methods is to learn a mapping from low-level features extracted from local image patches to the count~\cite{chan2009bayesian,ryan2009crowd}. These extracted features include foreground features, edge features, textures, and gradient features such as local binary pattern\;(LBP), and histogram oriented gradients\;(HOG). The regression approaches include linear regression~\cite{Paragios2001}, piece-wise linear regression~\cite{Chan2008Privacy}, ridge regression~\cite{Chen2012Feature}, and Gaussian process regression. Although these methods refine the previous detection-based ones, they ignore spatial distribution information of crowds. To utilize spatial distribution information, the method proposed by Lempitsky \textit{et al.}~\cite{lempitsky2010learning} regresses a density map rather than the crowd count. The method learns a linear mapping between local patch features and density maps, then estimates the total number of pedestrians via integrating over the whole density map. The method proposed by Pham \textit{et al.}~\cite{Pham2015} learns a non-linear mapping between local patch features and density maps by using random forest. Most recent regression-based methods are based on the density map.

\subsection{\textbf{CNN-based Methods}}
Benefit from  CNN's strong ability to learn representations, a variety of CNN-based methods have recently been introduced in crowd counting. As a pioneering work for crowd counting with CNN, the method proposed by Wang \textit{et al.}~\cite{wang2015deep} adopted multiple convolutional layers to extract features and sent these features into a fully connected layer to predict numbers in extremely dense crowds. Another work~\cite{zhang2015cross} pretrained a network for certain scenes and selected similar training data to fine-tune the pretrained network based on the perspective information. The main drawback is that the approach requires perspective information which is not always available.

Observing that the densities and appearances of image patches are of large variations, Zhang \textit{et al.}~\cite{Zhang2016Single} further proposed MCNN architecture for estimating the density map. In their work, different columns are explicitly designed for learning density variations across different feature resolutions. Despite different sizes of filters, it is difficult for different columns to recognize varying density crowds, and this lack of recognition results in some ineffective branches. Sindagi \textit{et al.}~\cite{Sindagi2017} proposed a multi-task framework to simultaneously predict density classification and generate the density map based on high-level prior information. They further proposed a five-branch contextual pyramid CNNs, short for CP-CNN~\cite{sindagi2017generating}, to incorporate contextual information of the crowd for achieving lower counting errors and high-quality density maps. However, CP-CNN has high computational complexity and cannot be applied in real-time scene analysis. Inspired by MCNN, the work proposed by Sam \textit{et al.}~\cite{sam2017switching} includes a Switch-CNN, where the switch classifier is trained to select the optimal regressor for one input patch. In the prediction phase, Switch-CNN can only use a column network that is consistent with the classification result of that patch, without incorporating all trained subnetworks. High-variation densities not only  exist at the whole image level, but also at the image patch level. Therefore, the single subnetwork has  limited recognition performance and cannot overcome the feature covariate shift problem. Kang \textit{et al.}~\cite{kang2018crowd} proposed a method of fusing multi-scale density predictions of corresponding multi-scale inputs, while Deb \textit{et al.}~\cite{deb2018aggregated} designed an aggregated multi-column dilated convolution network for perspective-free counting. However, none of these works consider local information.
 \begin{figure}[t]
 \centering
  \subfigure[$C = 50$, $D = 24.1$]{\includegraphics[height=0.38\linewidth,width=0.48\linewidth]{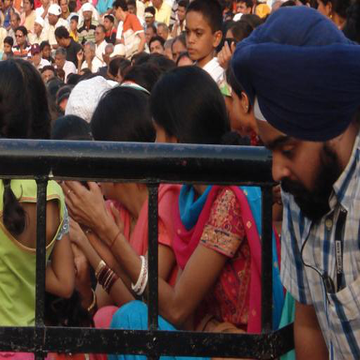}}
  \subfigure[$C = 50, D = 55.0$]{\includegraphics[height=0.38\linewidth,width=0.48\linewidth]{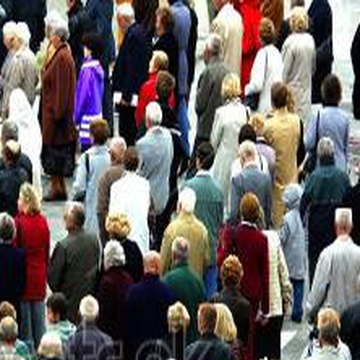}}\\
  \subfigure[$C = 78, D = 20.8$]{\includegraphics[height=0.38\linewidth,width=0.48\linewidth]{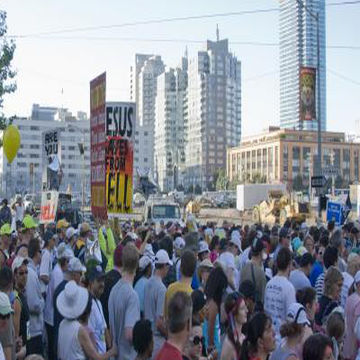}}
  \subfigure[$C = 241, D = 20.8$]{\includegraphics[height=0.38\linewidth,width=0.48\linewidth]{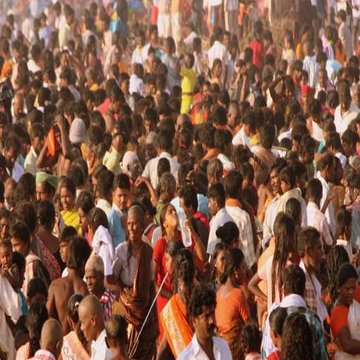}}
 \caption{The average distance between adjacent people is more reasonable for representing the dense degree of crowd compared with the number of people. Several instances of the SHA dataset~\cite{Zhang2016Single} are shown. $C$ is the number of people. $D$ is the dense degree of the crowd calculated by Eq.~\eqref{Equ1} (smaller is denser). The number of people is the same in figures (a) and (b), but (a) is denser than (b) and $D$ in (a) is smaller than that in (b). On the other hand, $D$ is the same in figures (c) and (d), but the number of people in (c) is far less than that in (d).}
 \label{fig4}
\end{figure}

To avoid the issues of ineffective branches and expensive computation in previous multi-column networks, Li \textit{et al.}~\cite{li2018csrnet} introduced a deeper single-column-based dilated convolutional network called CSRNet. Cao \textit{et al.}~\cite{Cao_2018_ECCV} developed an encoder-decoder-based scale aggregation network for crowd counting. Observing the importance of temporal information in counting crowds, a bi-directional ConvLSTM-based~\cite{Shi2015Convolutional} spatiotemporal model was proposed  for video crowd counting~\cite{Xiong2017Spatiotemporal}.

Most of the CNN-based methods count crowds by predicting a density map based on $l_2$ regression loss. However, $l_2$ is sensitive to outliers and blurs the density map. Shen \textit{et al.}~\cite{shen2018crowd} thus proposed a GANs-based method to generate high-quality density maps, and a strong regularization constraint was conducted on cross-scale crowd density estimation. In addition, Liu \textit{et al.}~\cite{liu2018decidenet} combined the detection-based and the regression-based method for dealing with density variation for crowd coutning. Shi \textit{et al.}~\cite{shi2018crowd} proposed a framework  which produced generalizable features by using deep negative correlation learning (NCL). Liu \textit{et al.}~\cite{Liu_2018_CVPR} leveraged unlabeled data to enhance the feature representation capability of the network. Inspired by image generation, the method proposed by Ranjan \textit{et al.}~\cite{Ranjan_2018_ECCV} is an iterative crowd counting framework which generates a low-quality density map first and then gradually evolves it to a high-quality density map. Note that these methods are inferior in achieving robust recognition performance in pan-density crowd counting, which is the problem we are tackling.

\section{\textbf{Our Approach}}\label{sec:met}

Our framework is illustrated in Figure~\ref{fig3}. The proposed PaDNet consists of four components: Feature Extraction Network\;(FEN), Density-Aware Network\;(DAN), Feature Enhancement Layer\;(FEL), and Feature Fusion Network\;(FFN). FEN extracts the low-level features of images. DAN contains multiple subnetworks pretrained on scenarios with different densities and is used to capture pan-density information. FEL captures the global and local features to learn an enhancement rate or a weight for each density-specific feature map. Finally, FFN aggregates all of the refined features to generate the final density map for counting the crowd. Below, we will describe our proposed PaDNet in details.

\subsection{\textbf{Feature Extraction Network\;(FEN)}}

A main difficulty in crowd counting is that the background and the density level have drastical variations in real-world scenes. To apply deep learning for such a situation, a sufficiently large training dataset is required. However, the largest existing training dataset only contains 1,201 images. As was done in many deep learning models~\cite{li2018csrnet,sindagi2017generating,shi2018crowd, shen2018crowd,Liu_2018_CVPR}, we use pretrained models to avoid overfitting. Note that because most of the popular backbones such as VGG-16~\cite{vgg}, ResNet~\cite{resnet}, and GoogLeNet~\cite{googlenet} are trained on the ImageNet~\cite{Russakovsky2015ImageNet}, which is a classification task, while crowd counting is a regression task, these backbones cannot be directly inserted into our model. Thus, we have to develop an alternative process.
The work of Yosinski \textit{et al.}~\cite{NIPS2014_5347} considers that the front-end of the network learns task-independent general features which are similar to Gabor filters and color blobs, and the back-end of the network learns task-specific features. Based on this consideration, we choose the first ten convolutional layers of a pretrained VGG-16 with Batch Normalization~\cite{BN} and ReLU as FEN.
Formally, the feature extraction process is as follows,
\begin{align}
 \hat{Y}_{\mbox{\footnotesize feat}} &= F_{\mbox{\footnotesize FEN}}(X)\;,
 \end{align}
where $X$ is an input image, $\hat{Y}_{\mbox{\footnotesize feat}}$ is the base feature and has 512 channels, and $F_{\mbox{\footnotesize FEN}}(\cdot)$ is the FEN.

\begin{algorithm}[ht!]
\caption{Training Phase}
\label{al1}
\hspace*{0.02in}{\bf Input:}
input crowd image patches dataset $S$\\
\hspace*{0.02in}{\bf Output:}
output the parameters $\Theta_{\mbox{\footnotesize PaDNet}}$ \\
\hspace*{0.02in}{\bf Init:}
Dividing the whole image patches $S$ into $N$ clusters $S_{1}, S_{2} ... S_{N}$ via K-means clustering algorithm.
\begin{algorithmic}
\For {$i$ = 1 to $epoch_{1}$}
 \For {$j$ = 1 to $N$}
 \State Training $j$th subnetwork with $S_{j}$ update $\Theta_{j}$
 \State $\mathcal{L}_{mse} = \frac {1} {\lVert S_j \rVert} \sum_{i=1}^{\lVert S_j \rVert} \lVert \hat{Y_i} - Z_{i}^{\mbox{\footnotesize GT}}\rVert_2^2$
 \State Saving the best state $\Theta_{j}$ of $j$th subnetwork
 \EndFor
\EndFor
\State Loading the best $\{\Theta{j}\}_1^N$ for \textsl{PaDNet}
\For {$i$ = 1 to $epoch_2$ }
 \State Training \textsl{PaDNet} with $S$ update $\Theta_{\mbox{\footnotesize PaDNet}}$ and don't freeze the $\Theta_{j}$.
 \State $\mathcal{L} = \frac {1} {\lVert S \rVert} \sum_{i=1}^{\lVert S \rVert} \lVert F_{\mbox{\footnotesize PaDNet}}(X_{i}) - Z_{i}^{\mbox{\footnotesize GT}}\rVert_2^2\ + \lambda \mathcal{L}_{\mbox{\footnotesize ce}}$
\EndFor
\State \Return $\Theta_{\mbox{\footnotesize PaDNet}}$
\State \textsl{Adam} is applied with learning rate at $10^{-5}$ and weight decay at $10^{-4}$
\end{algorithmic}
\end{algorithm}
\subsection{\textbf{Density-Aware Network\;(DAN)}}
The goal of DAN is to capture pan-density information. Therefore, each subnetwork in DAN is pretrained on scenarios with specific density so that it can recognize specific density crowds. However, determining the ground truth for an image's density level depends on human experiences. A straightforward way is to focus on the number of people in the image. Due to differences in crowd distributions, there exists scenarios where the number of people is the same while the density of the crowd is different. To address this issue, the work proposed by Sam \textit{et al.}~\cite{sam2017switching} suggests that using the average distance between adjacent heads is more effective than mere head count to represent the dense degree of the crowd. Therefore, we calculate the dense degree for an image patch as follows,
 \begin{align}
 D &= \frac {1}{P} \sum_{i=1}^P \sum_{j=1}^Q d_{ij}\;,
 \label{Equ1}
 \end{align}
where $P$ is the number of people in an image patch, $d_{ij}$ represents the distance between the $i$th subject and its $j$th nearest neighbor, and $Q$ is the calculated maximum number of nearest neighbors. Intuitively, the smaller the value of $D$, the denser the crowd. Examples are shown in Figure~\ref{fig4} indicates that the average distance is more reasonable to represent the dense degree of the crowd.

\begin{table}[ht!]
 \caption{DAN consists of multiple subnetworks. The number of subnetwork is related to classes of density level. The convolutional layers' parameters are denoted as ``Conv(kernel\_size, output\_channels).'' Every convolutional layer is followed by Batch Normalization~\cite{BN} and ReLU.}
 \renewcommand\arraystretch{1.9}
 \begin{tabular}{|p{1.7cm}<{\centering}|p{1.7cm}<{\centering}|p{1.7cm}<{\centering}|p{1.7cm}<{\centering}|}
 \hline
 Level-1 & Level-2 & Level-3 & Level-4 \\

 \hline
 \hline
 Conv(9, 384) & Conv(7, 256) & Conv(5, 128) & Conv(5, 128) \\
 \hline
  Conv(9, 256) & Conv(7, 128) & Conv(5, 64) & Conv(5, 64)\\
 \hline
 Conv(7, 128) & Conv(5, 64) & Conv(3, 32) & Conv(3, 32) \\
 \hline
 Conv(5, 64) & Conv(3, 32) & Conv(3, 16) & Conv(3, 16) \\
 \hline
 Conv(1, 1) & Conv(1, 1) & Conv(1, 1) & Conv(1, 1) \\
 \hline
\end{tabular}
 \centering
 \label{tab1}
\end{table}

In DAN, the number of subnetworks is the same as the number of  density categories. We design different network configurations from level-1 to level-4 subnetworks. The configurations are shown in Table~\ref{tab1}. The lower-level networks are used to recognize sparse crowds; the higher-level networks are used to recognize dense crowds. For sparse crowds, the distances between adjacent people are larger and the head sizes are typically larger than they are in dense crowds. Therefore, we use lower-level subnetworks with larger filters to recognize the sparser crowds and higher-level subnetworks with smaller filters to recognize denser crowds. As the density level increases, the sizes of the filters gradually become smaller. The filters of each subnetwork are pyramidal and become smaller for enhancing the multi-scale ability of the subnetwork. In addition, the lower subnetworks include more filters than the higher subnetworks do in each layer because the distribution of a dense scene is more uniform than it is in a sparse scene. The work of Li \textit{et al.}~\cite{li2018csrnet} suggests that too many pooling layers can reduce the spatial information of feature map. Therefore, there is no pooling layer in DAN. Formally, given the base feature $\hat{Y}_{\mbox{\footnotesize feat}}$ as input for DAN, each subnetwork generates a density-specific feature as follows,
  \begin{align}
    \hat{Y}_i &= F_{{\mbox{\footnotesize sub}}_i}(\hat{Y}_{\mbox{\footnotesize feat}})\;,
 \label{Equ2}
 \end{align}
where $\hat{Y}_i$ is the density-specific feature generated by the $i$th subnetwork, and $F_{{\mbox{\footnotesize sub}}_i}(\cdot)$ denotes the $i$th subnetwork of DAN.

\subsection{\textbf{Feature Enhancement Layer\;(FEL)}}
Although each subnetwork of DAN can recognize a specific density crowd, feature maps at varying levels of density should be weighted with different importance because of the nonuniform distributions of crowds. Therefore, we design a Feature Enhancement Layer\;(FEL) to assign different weights for varying feature maps. Specifically, these subnetworks of DAN generate density-specific feature maps, $\hat{Y}_{1}$, $\hat{Y}_{2}$, $...$ , $\hat{Y}_{n}$. We concatenate them as input for FEL. FEL consists of a Spatial Pyramid Pooling (SPP)~\cite {he2014spatial} and a Fully Connected (FC) layer. SPP performs three scales pooling operations for each $\hat{Y}_i$. The $i$th operation divides feature map into $i\times i$ regions with pooling to capture the global and local contextual features. Then the FC layer analyzes the contextual features to weight the importance for each $\hat{Y}_i$.  Formally, given several density-specific feature maps $\hat{Y}_{1}$, $\hat{Y}_{2}$, $...$ , $\hat{Y}_{n}$ as input for FEL as follows,
\begin{align}
    \boldsymbol{v} &= F_{\mbox{\footnotesize FEL}}(\hat{Y}_{1}, \hat{Y}_{2}, ... , \hat{Y}_{n})\;,
  \end{align}
where $F_{\mbox{\footnotesize FEL}}$ is FEL and $\boldsymbol{v}$ is the vector ($v_1$, $v_2$, $...$, $v_N$) generated by FEL, we weight the importance for each density-specific feature $\hat{Y}_i$  as follows,
\begin{align}
 \hat{Y}_i^{\prime} &= \hat{Y}_i \times (1 + w_i)\;,
 \label{Equ3}
 \end{align}
\begin{align}
    w_i &= \frac {\exp(v_{i})} {\sum_{j=1}^N \exp(v_{j})}\;,
\end{align}
where the number $1$ denotes that retaining the original feature of the $i$th subnetwork, and $w_i$ denotes the importance  for  this feature map. The cross-entropy loss is used to train FEL.

\subsection{\textbf{Feature Fusion Network (FFN)}}
In order to fuse the refined feature maps $\hat{Y^{\prime}_i}s$, we propose a sophisticated network, named Feature Fusion Network\;(FFN), to embed spatial context effectively and combine all feature maps for generating the final density map. FFN consists of Conv(7, 64), Conv(5, 32), Conv(3, 32), and Conv(1, 1). Every convolutional layer is followed by Batch Normalization and ReLU. Inspired by U-Net~\cite{unet} and DenseNet~\cite{densehuang2017}, skip connections can make up for the lost information and improve the performance. Before Conv(1, 1) layer, we further add a skip connection to concatenate $\hat{Y_i}s$.

The detail of the training procedure is shown in Algorithm~\ref{al1}. The loss function for training the PaDNet is given as follows,
 \begin{align}
 \mathcal{L} &= \mathcal{L}_{\mbox{\footnotesize mse}} + \lambda \mathcal{L}_{\mbox{\footnotesize ce}} \;, \\
 \mathcal{L}_{\mbox{\footnotesize mse}} &= \frac {1}{M} \sum_{i=1}^M \lVert F_{\mbox{\footnotesize PaDNet}}(X_{i}) - Z_{i}^{\mbox{\footnotesize GT}}\rVert_2^2\;,
 \end{align}
where $M$ is the number of the training samples and $\lambda$ is the weight factor of $\mathcal{L}_{\mbox{\footnotesize ce}}$ with the settings listed in Table~\ref{tab2}. The denser the crowd, the larger the $\lambda$. In a sparse crowd, the value of training loss $\mathcal{L}_{\mbox{\footnotesize mse}}$ is very small. Therefore, we set a small value for $\lambda$.  $F_{\mbox{\footnotesize PaDNet}}(X_{i})$ is the density map estimated by the PaDNet. $Z_{i}^{\mbox{\footnotesize GT}}$ is the $i$th ground truth.

\begin{table}[h]
 \caption{The parameter settings of $\lambda$ for different datasets.}
 \renewcommand\arraystretch{1.2}
 \begin{tabular}{p{3cm}<{\centering}|p{3cm}<{\centering}}
 \hline
 Dataset & $\lambda$ \\

 \hline
 \hline
 ShanghaiTech\_A\cite{Zhang2016Single} & \multirow{3}{*}{$1$} \\
 \cline{1-1}
 UCF\_CC\_50\cite{Idrees2013Multi} & \\
 \cline{1-1}
 UCF\_QNRF\cite{idrees2018composition} & \\
 \hline
 ShanghaiTech\_B\cite{Zhang2016Single} & $0.1$ \\
 \hline
 UCSD\cite{Chan2008Privacy} & $0.01$ \\
 \hline

\end{tabular}
 \centering

 \label{tab2}
\end{table}

\section{\textbf{Experiments}}\label{sec:exp}
We now evaluate PaDNet on four crowd counting benchmark datasets: the ShanghaiTech~\cite{Zhang2016Single}, the UCSD~\cite{Chan2008Privacy}, the UCF\_CC\_50~\cite{Idrees2013Multi}, and the UCF-QNRF~\cite{idrees2018composition}. We compare PaDNet with five state-of-the-art algorithms including D-ConvNet~\cite{shi2018crowd}, ACSCP~\cite{shen2018crowd}, ic-CNN~\cite{Ranjan_2018_ECCV}, SANet~\cite{Cao_2018_ECCV}, and CSRNet~\cite{li2018csrnet}. Further, we conduct extensive ablation experiments to analyze the effect of different components in PaDNet. We detail experimental settings and results below.

\subsection{\textbf{Data Preparation}}
We resize the training images to 720 $\times$ 720, and crop nine patches from each image. Four of them contain four quarters of the image without overlapping. The remaining five patches are randomly cropped from the image. By using horizontal flip for these patches, we can get 18 patches from each image. We calculate the dense degree $D$ using Eq. \eqref{Equ1} for every patch. Then the $K$-means algorithm is performed to cluster all image patches into $C$  subsets with varying density level. To avoid sample imbalance, we continue to randomly crop the patches from the original images to balance each subset  so that each category will have an equivalent number of patches. The setups of different datasets are listed in Table~\ref{tab3}. Note that the UCSD~\cite{Chan2008Privacy} is a sparse dataset, hence we set $Q$ to $2$.

\begin{table}[ht!]
 \caption{Q nearest neighbors are calculated for different datasets.}
 \renewcommand\arraystretch{1.2}
 \begin{tabular}{p{3cm}<{\centering}|p{3cm}<{\centering}}
 \hline
  Dataset & $Q$ Nearest Neighbors \\
  \hline
 \hline
  ShanghaiTech\cite{Zhang2016Single} & \multirow{3}{*}{$ 5$}  \\

  UCF\_CC\_50\cite{Idrees2013Multi} & \\

  UCF\_QNRF\cite{idrees2018composition} & \\

  \hline
  UCSD\cite{Chan2008Privacy} & $2$ \\
 \hline
\end{tabular}
 \centering
 \label{tab3}
\end{table}
The ground truth is generated by blurring the head annotations with a normalized Gaussian kernel (sum to one). Geometry-adaptive kernel used for generating the density map, as in~\cite{Zhang2016Single}, is defined as:
 \begin{align}
  F(x) = \sum_{i=1}^N \delta(x - x_i) \times G_{\sigma_{i}}(x), \mbox{with} \; \sigma_{i} = \beta \overline{d_{i}}\;,
 \end{align}
where $x_i$ is the position of $i$th head in the ground truth $\delta$ and $\overline{d_{i}}$ is the average distance of $K$ nearest neighbors. We convolve $\delta(x - x_i)$ with a Gaussian kernel with parameter $\sigma_{i}$. For the ShanghaiTech~\cite{Zhang2016Single}, the UCF\_CC\_50~\cite{Idrees2013Multi}, and the UCF-QNRF~\cite{idrees2018composition} datasets, we set $\beta$ to $0.3$. The UCSD dataset~\cite{Chan2008Privacy} does not satisfy the assumptions that the crowd is evenly distributed, so we set $\sigma$ of the density map to 3.

\subsection{\textbf{Evaluation Metrics}}
 The general evaluation metrics of crowd counting are mean absolute error\;(MAE) and root mean squared error\;(RMSE). Here MAE is defined as
  \begin{align}
  \mbox{MAE} &= \frac {1}{M} \sum_{i=1}^M \lvert {C_{X_{i}} - C_{X_{i}}^{\mbox{\footnotesize GT}}} \rvert\;,
  \end{align}
 and RMSE is defined as
 \begin{align}
 \mbox{RMSE} &= \sqrt{\frac{1}{M}\sum_{i=1}^M \left(C_{X_{i}} - C_{X_{i}}^{\mbox{\footnotesize GT}}\right)^{2}}\;,
 \end{align}
 where $M$ is the number of test samples, $C_{X_{i}}$ and $C_{X_{i}}^{\mbox{\footnotesize GT}}$ are the estimated number of people and the ground truth for the $i$th image, respectively. Moreover, the MAE and the RMSE reflect the algorithm's global accuracy and robustness.

 However,  MAE and RMSE cannot be used to evaluate local performance. The GAME metric~\cite{Guerrero2015Extremely} that has been used in vehicle counting to evaluate local estimation has some similarity. But it just covers a limited range of scales and does not adequately reflect the robustness for pan-density crowd counting. Therefore, we expand MSE and RMSE to patch mean absolute error\;(PMAE) and patch root mean squared error\;(PRMSE), respectively to accommodate our needs.
\begin{align}
  \mbox{PMAE} &= \frac {1}{n \times M} \sum_{i=1}^{n \times M} \lvert {C_{X_{i}} - C_{X_{i}}^{\mbox{\footnotesize GT}}} \rvert\;,
  \label{PMAE}
  \end{align}
 \begin{align}
 \mbox{PRMSE} &= \sqrt{\frac{1}{n \times M} \sum_{i=1}^{n \times M}\left(C_{X_{i}} - C_{X_{i}}^{\mbox{\footnotesize GT}}\right)^{2}}\;.
 \label{PRMSE}
 \end{align}
Specifically, we split each image into $n$ patches of same size  without overlapping and calculate the MAE and RMSE of the patches. That is, PMAE and PRMSE are able to completely reflect the algorithm's local accuracy and robustness. Note that when $n$ equals to 1, PMAE and PRMSE will degenerate into MAE and RMSE, respectively.
\begin{figure*}[ht!]
 \centering
  \includegraphics[width=\linewidth, clip=true, trim=0 100 0 0]{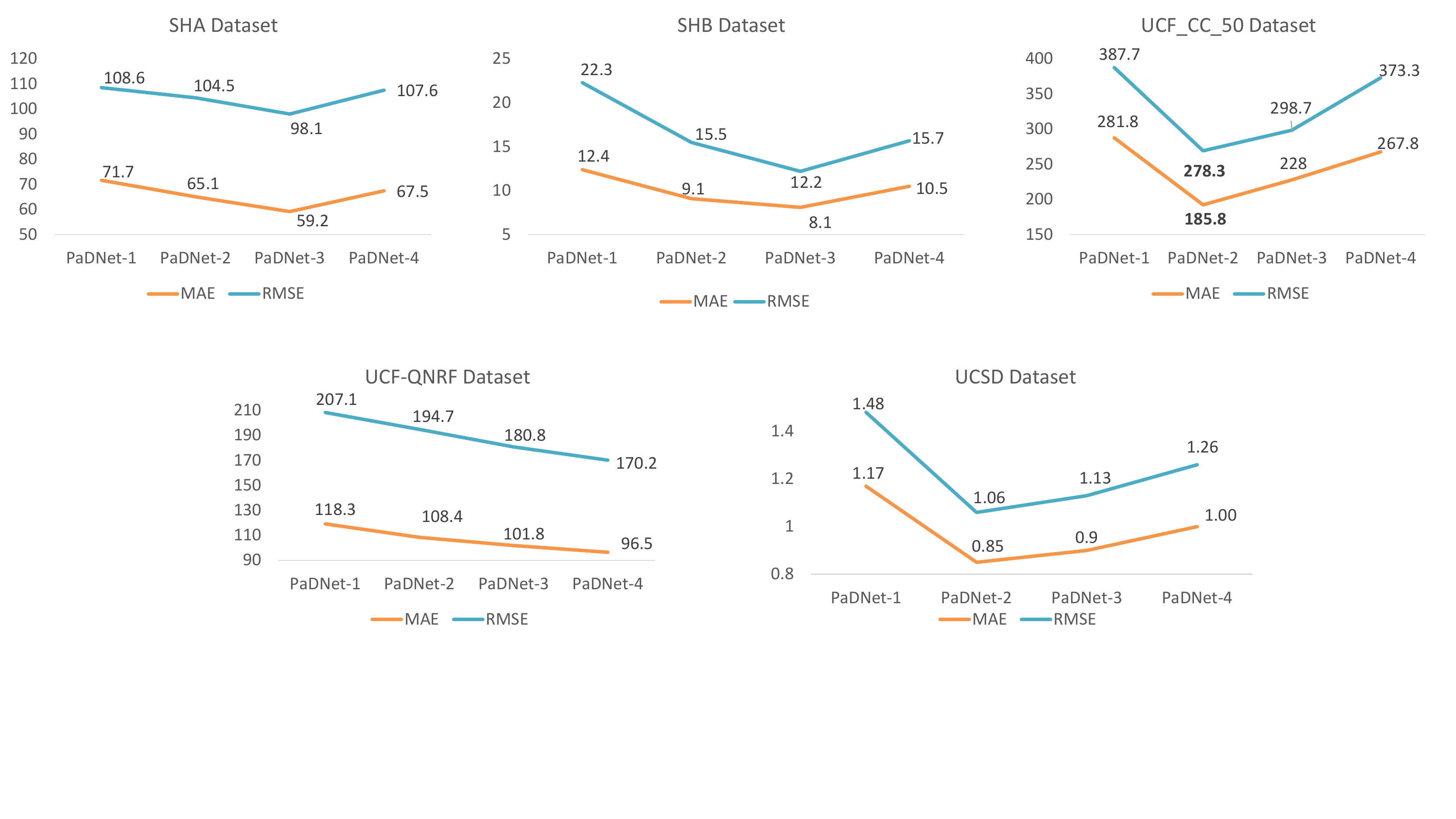}
 \caption{We conduct experiments on all datasets to analyze the effect of density level division. PaDNet-N indicates that we divide the dataset into $N$ classes, and PaDNet has $N$ subnetworks. PaDNet-2 achieves the best recognition performance on the UCSD and UCF\_CC\_50 datasets. PaDNet-3 has superior recognition performance on the ShanghaiTech dataset. PaDNet-4 performs the best on the UCF-QNRF dataset.}
 \label{fig7}
\end{figure*}
\subsection{\textbf{Datasets and Comparisons}}
\subsubsection{\textbf{The ShanghaiTech dataset}}
 This dataset contains 1,198 annotated images from a total of 330,165 people, each of which is annotated at the center of the head. The dataset is divided into Part\_A and Part\_B. Part\_A contains 482 images randomly crawled from the Internet. The training set has 300 images and the testing set has 182 images. Part\_B contains 716 images
  \begin{table}[t!]
 \caption{Comparison on the ShanghaiTech dataset}
 \renewcommand\arraystretch{1.2}
 \begin{tabular}{p{2.1cm}<{\centering}|p{0.8cm}<{\centering}|p{0.8cm}<{\centering}|p{0.8cm}<{\centering}|p{0.8cm}<{\centering}}
 \hline
 & \multicolumn{2}{c|}{Part\_A} & \multicolumn{2}{c}{Part\_B} \\
 \hline
  Method & MAE & RMSE & MAE & RMSE \\
 \hline
 \hline
  Zhang et al.~\cite{zhang2015cross} & 181.8 & 277.7 & 32.0& 49.8 \\

  MCNN~\cite{Zhang2016Single} & 110.2 & 173.2 & 26.4 & 41.3  \\

  Switch-CNN~\cite{sam2017switching} & 90.4 & 135.0 & 21.6 & 33.4 \\

  CP-CNN~\cite{sindagi2017generating} & 73.6 & 106.4 & 20.1 & 30.1 \\

  Liu et al.~\cite{Liu_2018_CVPR} & 73.6 & 112.0 & 13.7 & 21.4 \\

  IG-CNN~\cite{Sam_2018_CVPR} & 72.5 & 118.2 & 13.6 & 21.1 \\

  D-ConvNet~\cite{shi2018crowd} & 73.5 & 112.3 & 18.7 & 26.0 \\

  ACSCP~\cite{shen2018crowd} & 75.7 & 102.7 & 17.2 & 27.4 \\

  ic-CNN~\cite{Ranjan_2018_ECCV} & 68.5 & 116.2 & 10.7 & 16.0 \\

  CSRNet~\cite{li2018csrnet} & 68.2 & 115.0 & 10.6 & 16.0 \\

  SANet~\cite{Cao_2018_ECCV} & 67.0 & 104.5 & 8.4 & 13.6 \\

  {\bf Ours} & \textbf{59.2} & \textbf{98.1} & \textbf{8.1} & \textbf{12.2} \\
 \hline

 \end{tabular}
 \centering

 \label{tab4}
 \end{table}
    \begin{table}[ht!]
 \caption{Comparison on the UCF\_CC\_50 dataset}
 \renewcommand\arraystretch{1.2}
 \begin{tabular}{p{3.2cm}<{\centering}|p{1.3cm}<{\centering}|p{1.3cm}<{\centering}}
 \hline
  Method & MAE & RMSE \\
 \hline
 \hline
  Zhang et al.~\cite{zhang2015cross} & 467.0 & 498.5 \\

  MCNN~\cite{Zhang2016Single} & 377.6 & 509.1  \\

  Switch-CNN~\cite{sam2017switching} & 318.1 & 439.2  \\

  CP-CNN~\cite{sindagi2017generating} & 295.8 & 320.9 \\

  Liu et al.~\cite{Liu_2018_CVPR} & 337.6 & 434.3 \\

  IG-CNN~\cite{Sam_2018_CVPR} & 291.4 & 349.4 \\

  D-ConvNet~\cite{shi2018crowd} & 288.4 & 404.7 \\

  ACSCP~\cite{shen2018crowd} & 291.0 & 404.6 \\

  ic-CNN~\cite{Ranjan_2018_ECCV} & 260.9 & 365.5\\

  CSRNet~\cite{li2018csrnet} & 266.1 & 397.5 \\

  SANet~\cite{Cao_2018_ECCV} & 258.4 & 334.9 \\

  {\bf Ours} & \textbf{185.8} & \textbf{278.3} \\
 \hline

 \end{tabular}
 \centering
 \label{tab5}

 \end{table}
  taken from the busy streets of the metropolitan areas in Shanghai. The training set has 400 images, and the testing set has 316 images. The density of Part\_A is higher than Part\_B, and the density varies significantly. We test the performance of PaDNet on Part\_A and Part\_B as the other approaches did and report the best performance in Table~\ref{tab4}. PaDNet achieves the best performance among all approaches.
Specifically, it has an 11.6\% MAE and a 6.1\% RMSE improvement for the Part\_A dataset compared with the second-best approach, SANet~\cite{Cao_2018_ECCV}.

\subsubsection{\textbf{The UCF\_CC\_50 dataset}}
 This is an extremely dense crowd dataset. It contains 50 images of different resolutions with counts ranging from 94 to 4,543 with an average of 1,280 individuals in each image. The training set only has 40 images and the testing set only has 10 images. To more accurately verify
 \begin{table}[t!]
 \caption{Comparison on the UCSD dataset}
 \renewcommand\arraystretch{1.2}
 \begin{tabular}{p{3.2cm}<{\centering}|p{1.3cm}<{\centering}|p{1.3cm}<{\centering}}
 \hline
  Method & MAE & RMSE \\
 \hline
 \hline
  Zhang et al.~\cite{zhang2015cross} & 1.60 & 3.31 \\

  MCNN~\cite{Zhang2016Single} & 1.07 & 1.35  \\

  Switch-CNN~\cite{sam2017switching} & 1.62 & 2.10  \\

  ACSCP~\cite{shen2018crowd} & 1.04 & 1.35 \\

  Huang et al.~\cite{huangbody} & 1.00 & 1.40 \\

  CSRNet~\cite{li2018csrnet} & 1.16 & 1.47 \\

  SANet~\cite{Cao_2018_ECCV} & {1.02} & {1.29}\\

  {\bf Ours}& \textbf{0.85} & \textbf{1.06} \\
 \hline

 \end{tabular}
 \centering
 \label{tab6}
\end{table}
   \begin{table}[t!]
 \caption{Comparison on the UCF-QNRF dataset}
 \renewcommand\arraystretch{1.2}
 \begin{tabular}{p{3.2cm}<{\centering}|p{1.3cm}<{\centering}|p{1.3cm}<{\centering}}
 \hline
  Method & MAE & RMSE \\
 \hline
 \hline
  Idrees et al.~\cite{Idrees2013Multi}& 315.0 & 508.0 \\

 Encoder-Decoder~\cite{segnet} & 270.0 & 478.0 \\

  CMTL~\cite{Sindagi2017} & 252.0 & 514.0 \\

  ResNet101~\cite{resnet} & 190.0 & 277.0 \\

  DenseNet201~\cite{densehuang2017} & 163.0 & 226.0 \\

  MCNN~\cite{Zhang2016Single} & 277.0 & 426.0  \\

  Switch-CNN~\cite{sam2017switching} & 228.0 & 445.0  \\

  Idrees et al.~\cite{idrees2018composition} & 132.0 & 191.0 \\

  {\bf Ours} & \textbf{96.5} & \textbf{170.2} \\
 \hline

 \end{tabular}
 \centering
 \label{tab7}

 \end{table}
the performance of PaDNet, we adopt a 5-fold cross-validation following the standard setting in~\cite{Idrees2013Multi}. Experiments shown in Table~\ref{tab5} indicate that PaDNet achieves a 28.1\% MAE improvement compared with SANet,
 and a 13.3\% RMSE improvement compared with CP-CNN. These results indicate that PaDNet is suitable for extremely dense scenes.
 \begin{figure*}[ht!]
 \centering

  \subfigure[Original]{\includegraphics[height=0.20\linewidth]{figures/IMG_134.jpg}}
  \subfigure[Ground Truth Count: 162]{\includegraphics[height=0.20\linewidth]{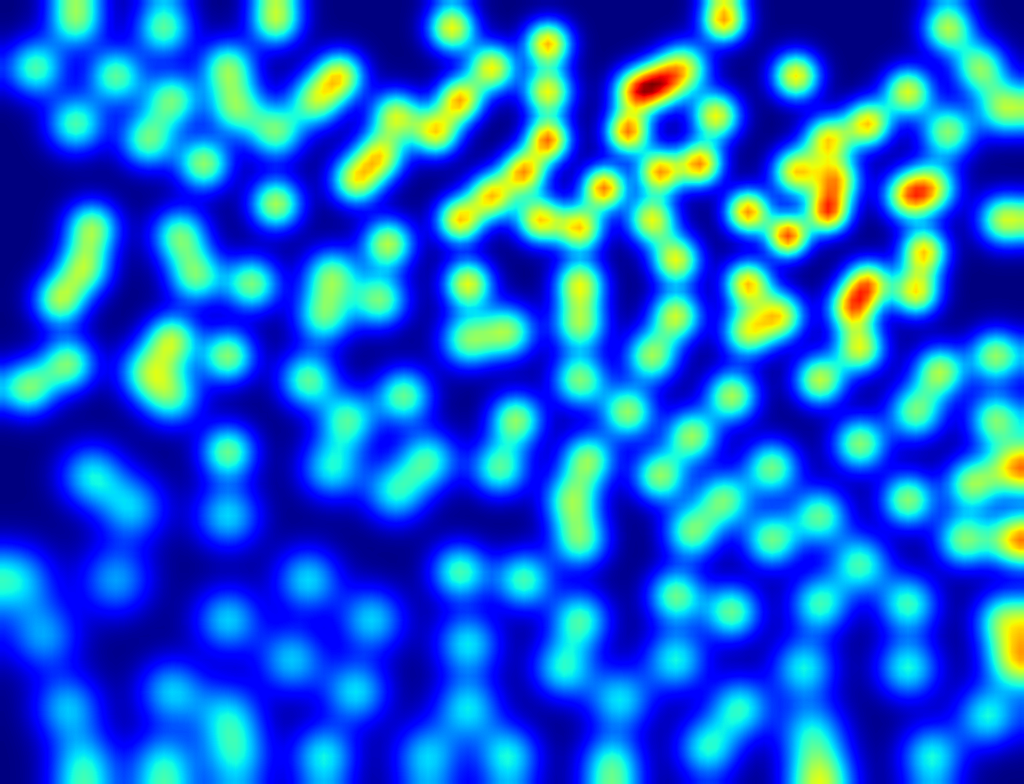}}
  \subfigure[PaDNet-1 Count: 74.8]{\includegraphics[height=0.20\linewidth]{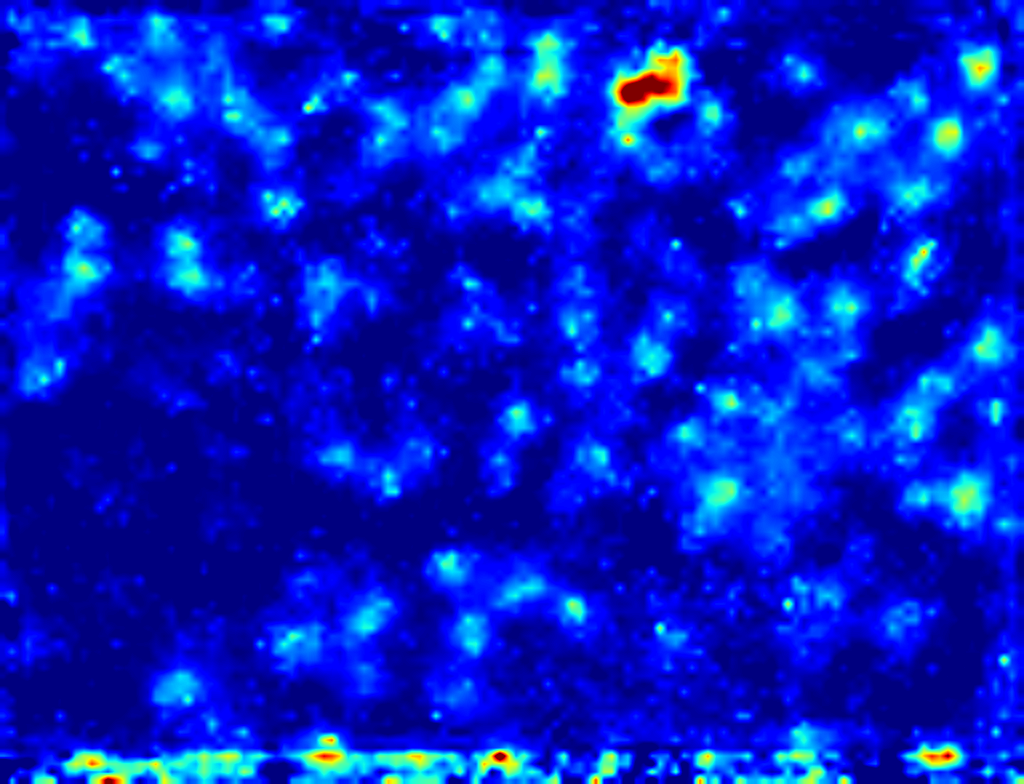}}
  \subfigure[PaDNet-2 Count: 127.2]{\includegraphics[height=0.20\linewidth]{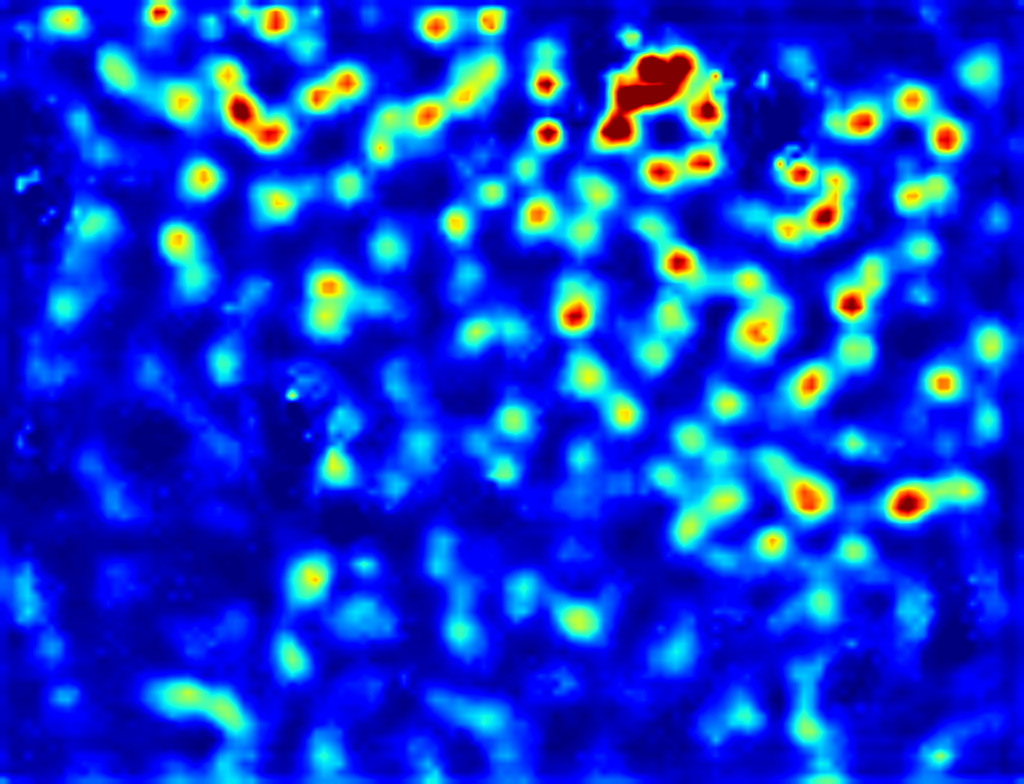}}
  \subfigure[PaDNet-3 Count: 136.5]{\includegraphics[height=0.20\linewidth]{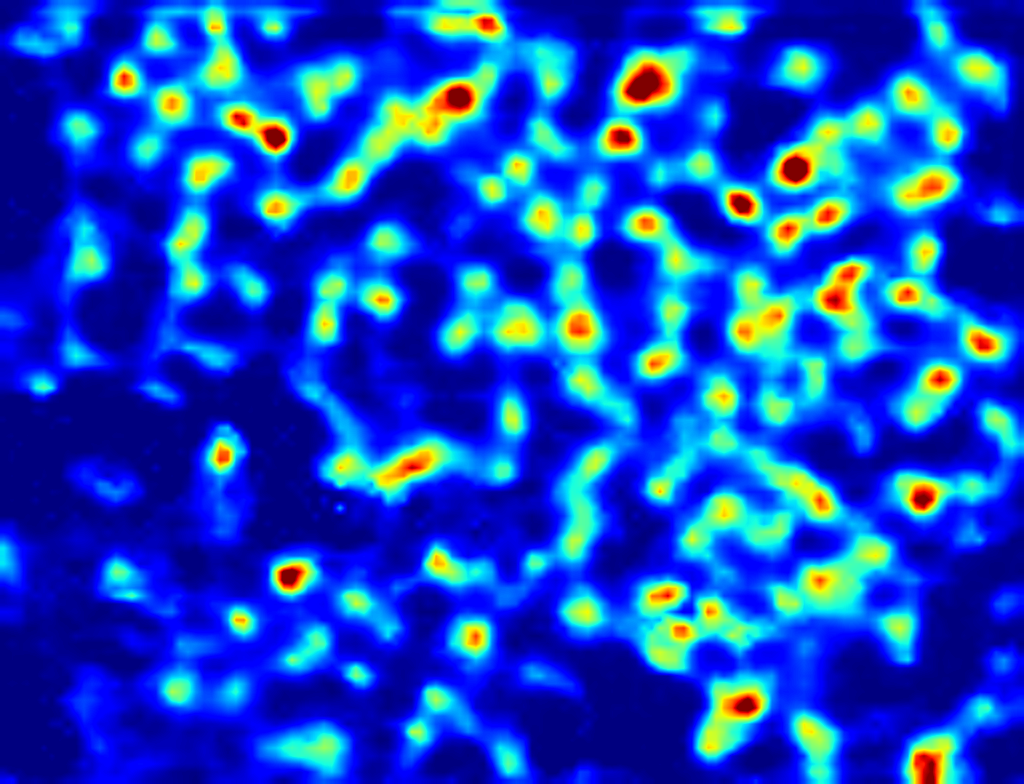}}
  \subfigure[PaDNet-4 Count: 75.3]{\includegraphics[height=0.20\linewidth]{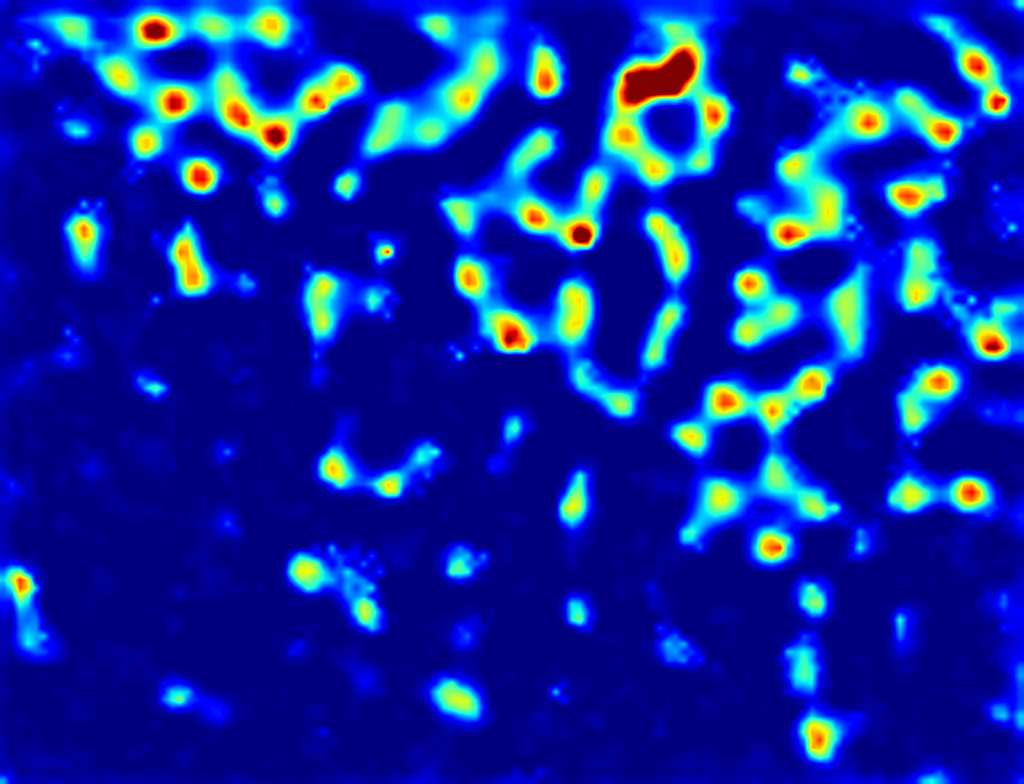}}
 \caption{An example result of the SHA dataset. (b) shows the ground truth density map. (c)-(f) are density maps generated by PaDNet-1, PaDNet-2, PaDNet-3 and PaDNet-4, respectively.}
 \label{fig8}
\end{figure*}

\subsubsection{\textbf{The UCSD dataset}}
The UCSD dataset~\cite{Chan2008Privacy} is a sparse density dataset that is a 2,000-frame video dataset chosen from one surveillance camera on the UCSD campus. The ROI and the perspective map are provided in the dataset. The resolution of each image is 238 $\times$ 158, and the crowd count in each image varies from 11 to 46. As Chan \textit{et al.} did, we use frames from 601 to 1,400 as the training set and the remaining frames for testing. All the frames and density maps are masked with ROI. The results are listed in Table~\ref{tab6}. Our method achieves superior performance both on a highly dense crowd dataset and a sparse crowd dataset. Our method has a 15.0\% MAE and a 17.8\% RMSE improvement for the UCSD dataset compared with the second-best approach, SANet and Huang {\it et al.}'s method~\cite{huangbody}.

\subsubsection{\textbf{The UCF-QNRF dataset}}
We further evaluate the recognition performance of PaDNet on the UCF-QNRF dataset~\cite{idrees2018composition}, which is the newest and largest crowd dataset. The UCF-QNRF contains 1.25 million humans marked with dot annotations and consists of 1,535 crowd images with wider variety of scenes containing the most diverse set of viewpoints, densities, and lighting variations. The minimum and the maximum counts are 49 and 12,865, respectively. Meanwhile, the median and the mean counts are 425 and 815.4, respectively. We use 1,201 images for training set and  334 images for testing, following~\cite{idrees2018composition}. The results are listed in Table~\ref{tab7}. PaDNet obtains the lowest MAE performance, and a 26.9\% MAE improvement compared with the second-best approach, {\it i.e.}, Idrees {\it et al.}~\cite{idrees2018composition}.
\subsection{\textbf{Algorithmic Studies}}
We explore PaDNet from three aspects: 1) the effect of density level, 2) the effects of different components in PaDNet, and 3) the performance of PaDNet in pan-density crowd counting.
\begin{figure*}[htb]
 \centering

  \subfigure[Original]{\includegraphics[height=0.14\linewidth]{figures/IMG_88.jpg}}
  \subfigure[GT Count: 86]{\includegraphics[height=0.14\linewidth]{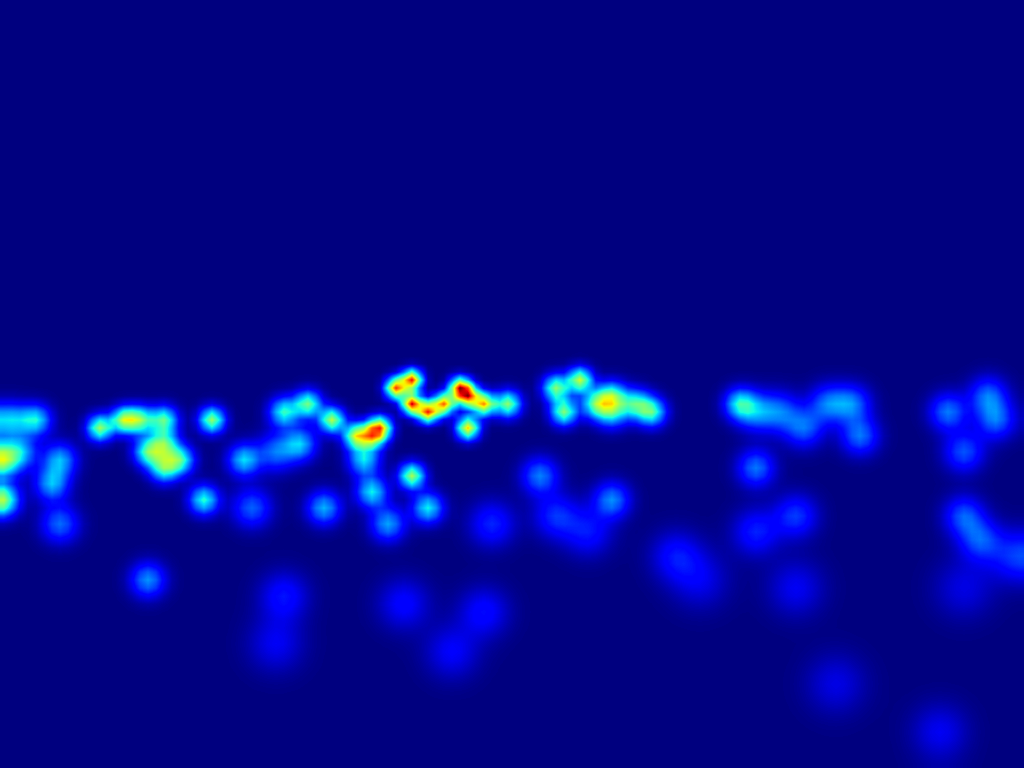}}
  \subfigure[Est Count: 107.6]{\includegraphics[height=0.14\linewidth]{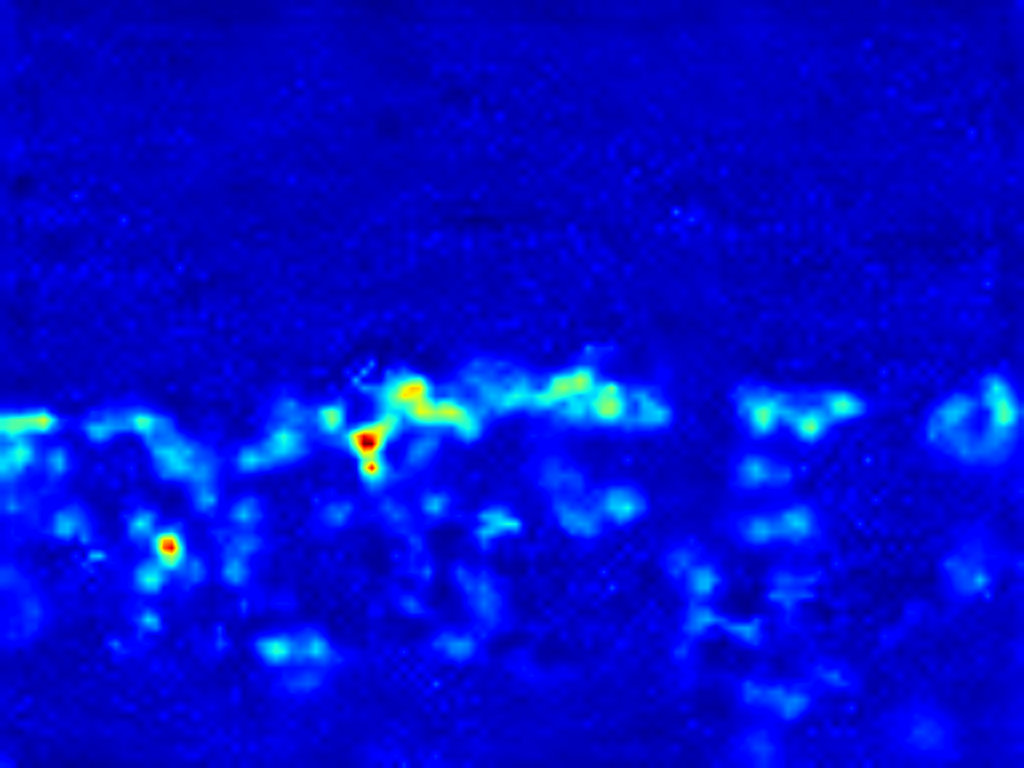}}
  \subfigure[Est Count: 80.2]{\includegraphics[height=0.14\linewidth]{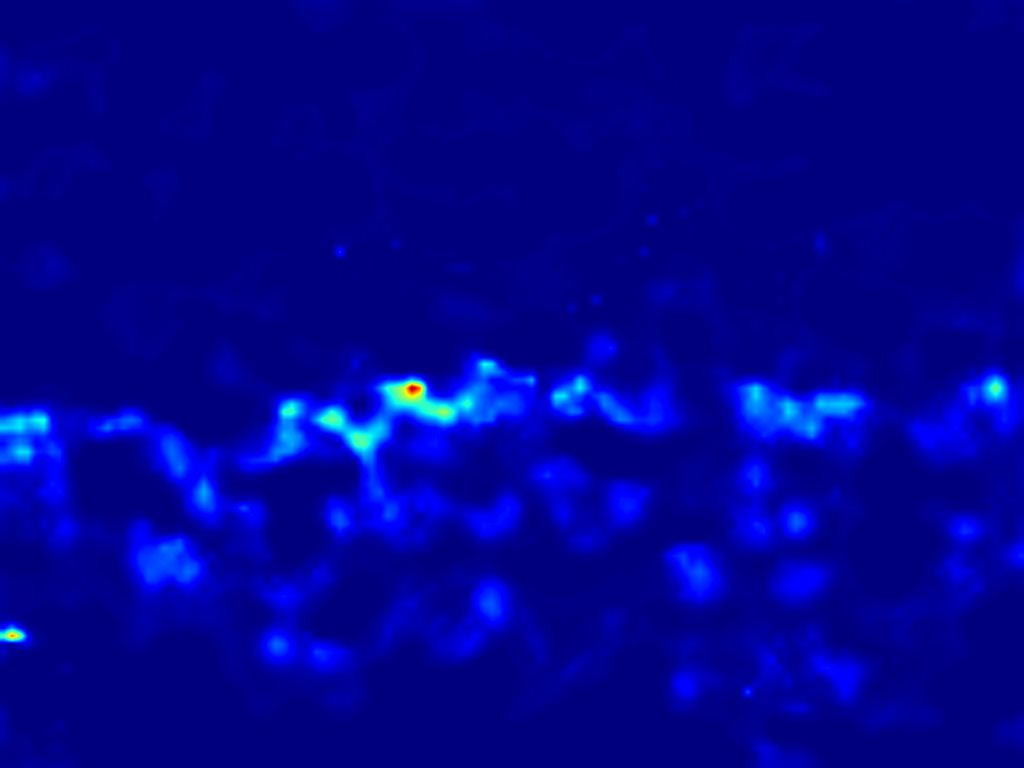}}
  \subfigure[Est Count: 83.0]{\includegraphics[height=0.14\linewidth]{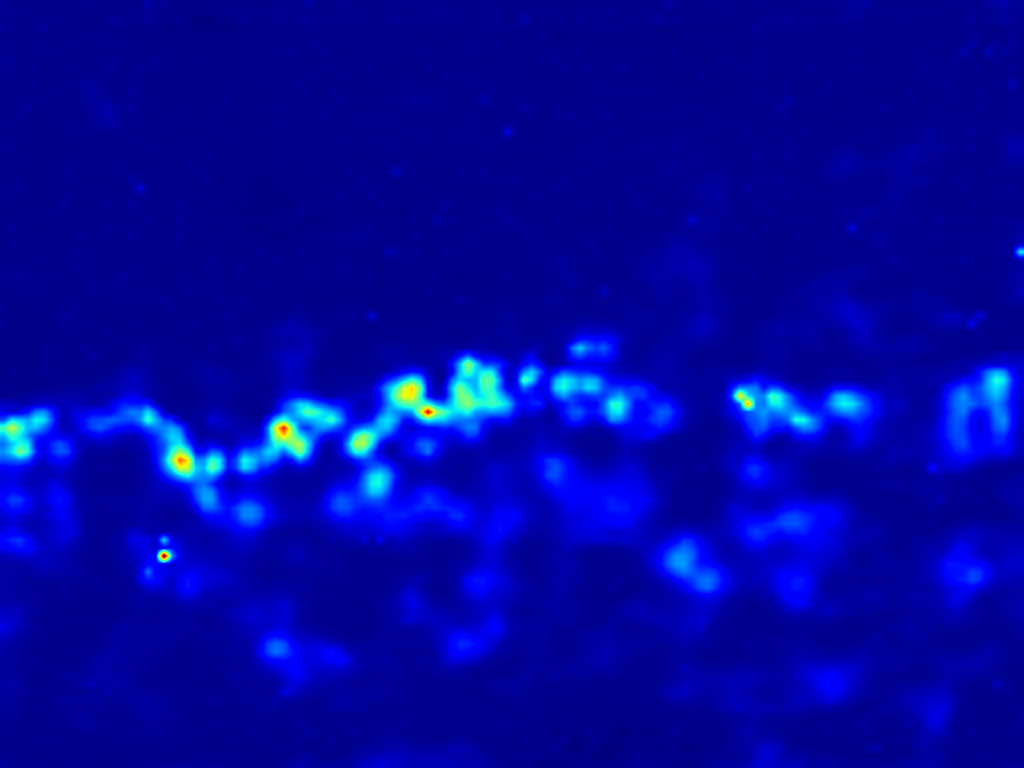}}
 \caption{An example result of the SHA dataset~\cite{Zhang2016Single}. The density maps are generated by different configuration of PaDNet. (b) shows the ground truth. (c) is the result of the PaDNet without FEL and Skip Connection (SC). (d) is the result of PaDNet only without SC. (e) is the result of PaDNet.}
 \label{fig9}
\end{figure*}
\begin{table*}[htb]
 \caption{The PMAE and PRMSE of PaDNet compare with CSRNet and MCNN.}
 \renewcommand\arraystretch{1.2}
 \begin{tabular}{p{2.6cm}<{\centering}|p{1.27cm}<{\centering}|p{1.27cm}<{\centering}|p{1.27cm}<{\centering}|p{1.27cm}<{\centering}|p{1.27cm}<{\centering}|p{1.27cm}<{\centering}|p{1.27cm}<{\centering}|p{1.27cm}<{\centering}}
 \hline
 \multirow{2}*{Methods} & \multicolumn{2}{c|}{n = 1} & \multicolumn{2}{c|}{n = 4} & \multicolumn{2}{c|}{n = 9} & \multicolumn{2}{c}{n = 16} \\

 \cline{2-9}
 ~ & PMAE & PRMSE & PMAE & PRMSE & PMAE & PRMSE & PMAE & PRMSE \\
 \hline
 \hline
 MCNN~\cite{Zhang2016Single} & 112.8 & 173.0 & 34.6 & 58.4 & 17.1 & 30.3 & 10.1 & 19.1\\

 CSRNet~\cite{li2018csrnet} & 68.8 & 107.8 & 19.8 & 37.3 & 9.6 & 19.9 & 5.7 & 13.2\\

 PaDNet w/o FEL\&SC & 65.0 & 103.2 & 20.6 & 38.5 & 10.6 & 21.6 & 6.3 & 14.1\\

 PaDNet w/o SC & 60.4 & 100.8 &18.3 & 35.8 & 9.1 & 19.3 &5.5 & 12.7\\

 PaDNet  & \textbf{59.2} & \textbf{98.1} &\textbf{17.9} & \textbf{35.4} & \textbf{8.8} & \textbf{19.1} &\textbf{5.3} & \textbf{12.7}\\

 \hline

 \end{tabular}
 \centering
 \label{tab8}
 \end{table*}

\subsubsection{\textbf{The effect of density level}}
The setting of the density level affects data preprocessing and the number of subnetworks, and this setting relates to  experimental results. The experimental results are shown in Figure~\ref{fig7}, where $N$ in the name PaDNet-$N$ indicates that we divide the dataset into $N$ classes and PaDNet has $N$ subnetworks. Specifically, when $N$ equals to 1, PaDNet does not have FEL or FFN. As seen in Figure~\ref{fig7}, PaDNet-2 achieves the best recognition performance on the UCSD and UCF\_CC\_50 datasets. PaDNet-3 has superior performance on the ShanghaiTech dataset. PaDNet-4 performs the best on the UCF-QNRF dataset.
\begin{table}[pt!]
  \caption{The effects of different components in PaDNet on the SHA dataset.}
 \renewcommand\arraystretch{1.2}
 \begin{tabular}{p{3.cm}<{\centering}|p{1.3cm}<{\centering}|p{1.3cm}<{\centering}}
 \hline

  Method & MAE & RMSE \\
\hline
\hline
 PaDNet w/o FEL\&SC & 65.0 & 103.2 \\

 PaDNet w/o SC & 60.4 & 100.8 \\

 PaDNet & \textbf{59.2} & \textbf{98.1} \\
\hline

 \end{tabular}
 \centering
 \label{tab10}
 \end{table}
Intuitively, different datasets should have been adapted to different number of subnetworks. On the other hand, the number of subnetworks should fit the distribution of the dataset. For examples, the crowd count in each image varies from 11 to 46 in the UCSD dataset, and the UCF\_CC\_50 dataset has an average of 1,280 persons in each image. Based on our experiments, we notice that when the division of density level is 2, the corresponding performance is the best because of the micro-variation density in the UCSD and UCF\_CC\_50 datasets.

For the ShanghaiTech dataset, PaDNet-3 performs better than PaDNet-2 because of the high-variation density in the ShanghaiTech dataset. Note that PaDNet-4 performs worse than PaDNet-3 on this dataset.
In general, we argue that the more abundant the training data for each subnetwork, the stronger the generalization ability of the subnetwork. Since PaDNet-1 only has one subnetwork, it is difficult to cover all the distributions of crowd. As the number of subnetworks increases, the amount of training data for each subnetwork decreases. While each subnetwork makes it easier to cover the distribution, it also reduces generalization ability to some extent. For the UCF-QNRF dataset which has a greater density variation, the minimum and the maximum counts are 49 and 12,865, respectively. It has 1,201 original images for training. Thus, we divide the UCF-QNRF dataset into 4 levels and PaDNet-4 achieves the best recognition performance.

In order to comprehend this reason intuitively, the density maps generated by PaDNet-1, PaDNet-2. PaDNet-3 and PaDNet-4
are shown in Figure~\ref{fig8}.
The density map generated by PaDNet-1 is slightly blurred and PaDNet-1 cannot recognize different density crowds. As the increase of subnetworks, the recognition abilities of PaDNet-2 and PaDNet-3 become gradually stronger. The generated density map achieves higher quality, and the estimated count is more precise. For PaDNet-4, although it has a clear density map, the estimated count is biased. As the generalization ability of each subnetwork weakens, which is caused by overfitting, PaDNet-4 cannot accurately recognize the bottom-left corner of the image.
\begin{table*}[ht]
  \caption{The PMAE and PRMSE of PaDNet-N compare with CSRNet and MCNN on more datasets.}
 \renewcommand\arraystretch{1.2}
 \begin{tabular}{p{2.2cm}<{\centering}|p{2.2cm}<{\centering}|p{1.0cm}<{\centering}|p{1.0cm}<{\centering}|p{1.0cm}<{\centering}|p{1.0cm}<{\centering}|p{1.0cm}<{\centering}|p{1.0cm}<{\centering}|p{1.0cm}<{\centering}|p{1.0cm}<{\centering}}
 \hline
 \multirow{2}*{Datasets} & \multirow{2}*{Methods} &\multicolumn{2}{c|}{n = 1} & \multicolumn{2}{c|}{n = 4} & \multicolumn{2}{c|}{n = 9} & \multicolumn{2}{c}{n = 16} \\
 \cline{3-10}
  ~ & & PMAE & PRMSE & PMAE & PRMSE & PMAE & PRMSE & PMAE & PRMSE \\
 \hline
 \hline
 \multirow{6}*{SHA~\cite{Zhang2016Single}}& MCNN~\cite{Zhang2016Single} & 112.8 & 173.0 & 34.6 & 58.4 & 17.1 & 30.3 & 10.1 & 19.1\\

    ~ & CSRNet~\cite{li2018csrnet} & 68.8 & 107.8 & 19.8 & 37.3 & 9.6 & 19.9 & 5.7 & 13.2\\

    ~& PaDNet-1 & 71.1 & 108.6 & 20.7 & 36.3 & 10.1 & 20.0 & 6.1 & 12.8\\

    ~& PaDNet-2 & 65.1 & 104.5 & 19.7 & 36.7 & 9.8 & 20.0 & 5.8 & 12.8\\

    ~& PaDNet-3  & \textbf{59.2} & \textbf{98.1} &\textbf{17.9} & \textbf{35.4} & \textbf{8.8} & \textbf{19.1} &\textbf{5.3} & \textbf{12.7}\\
    ~& PaDNet-4  & 67.5 & 107.6 & 19.4 & 37.1 & 9.4 & 20.2 & 5.6 & 12.9 \\

 \hline
 \multirow{6}*{SHB~\cite{Zhang2016Single}}& MCNN & 26.6 & 43.3 & 9.1 & 17.5 & 4.8 & 10.1 & 3.1 & 7.1\\

    ~ & CSRNet & 9.8 & 16.1 & 3.3 & 7.1 & 1.8 & 4.4 & 1.2 & 3.2\\

    ~& PaDNet-1 & 12.4 & 22.3 & 4.0 & 9.4 & 2.1 & 5.5 & 1.3 & 4.0\\

    ~& PaDNet-2 & 9.1 & 15.5 & 3.0 & 6.7 & 1.6 & 4.1 & 1.1 & 3.0\\

    ~& PaDNet-3  & \textbf{8.1} & \textbf{12.2} &\textbf{3.0} & \textbf{5.7} & \textbf{1.6} & \textbf{3.4} &\textbf{1.1} & \textbf{3.0}\\
    ~& PaDNet-4  & 10.5 & 15.8 & 3.6 & 7.0 & 1.9 & 4.2 & 1.3 & 3.1\\

 \hline
 \multirow{6}*{UCF\_CC\_50~\cite{Idrees2013Multi}}& MCNN & 378.1 & 504.3 & 114.1 & 179.4 & 53.6 & 84.8 & 32.3 & 54.0\\

    ~ & CSRNet & 256.4 & 355.2 & 73.7 & 110.7 & 36.5 & 60.3 & 21.6 & 36.1\\

    ~& PaDNet-1 & 281.8 & 387.7 &   82.3 & 121.9 & 39.6 & 64.7 & 23.7 & 39.4\\

    ~& PaDNet-2 & \textbf{185.8} & \textbf{278.3} & \textbf{65.9} & \textbf{93.3} & \textbf{33.4} & \textbf{53.6} &\textbf{20.2} & \textbf{32.2}\\

    ~& PaDNet-3  & 228.0 & 298.7 & 74.2 & 104.2 & 35.4 & 60.4 & 21.6 & 35.6\\
    ~& PaDNet-4  & 267.8 & 373.3 & 78.4 & 121.8 & 36.6 & 61.4 & 22.2 & 38.0\\

 \hline
 \multirow{6}*{UCF-QNRF~\cite{idrees2018composition}}& MCNN & 273.3 & 408.0 & 78.3 & 134.1 & 37.1 & 68.4 & 22.0 & 43.0\\

    ~ & CSRNet & 114.6 & 208.4 & 32.0 & 69.1 & 15.7 & 37.9 & 9.3 & 23.6\\

    ~& PaDNet-1 & 118.3 & 207.1 & 33.3 & 70.4 & 15.9 & 38.6 & 9.6 & 24.6\\

    ~& PaDNet-2 & 108.4 & 194.7 & 31.6 & 71.7 & 15.6 & 39.8 &9.5 & 25.1\\

    ~& PaDNet-3  & 101.8 & 180.8 & 30.4 & 65.4 & 15.4 & 37.7 & 9.3 & 23.7 \\
    ~& PaDNet-4  & \textbf{96.5} & \textbf{170.2} &\textbf{28.7} & \textbf{62.8} & \textbf{14.5} & \textbf{35.7} &\textbf{8.9} & \textbf{22.7}\\

 \hline

 \end{tabular}
 \centering
 \label{tab9}
 \end{table*}

\subsubsection{\textbf{Effects of different components}}
 We analyze the effects of different components of PaDNet in three aspects: (i) the effects of FEL and Skip Connection, (ii) the effect of DAN, and (iii) the effect of weighting strategy for feature maps.

\textbf{The effects of FEL and Skip Connection.} We conduct the experiments on the ShanghaiTech Part\_A\;(SHA) dataset to analyze the effects of FEL and Skip Connection\;(SC) in PaDNet-3. The results are listed in Table~\ref{tab10}.

The first method is the baseline of PaDNet-3 and does not have FEL or SC. In the second method, only FEL is introduced to analyze the effect of FEL. In the third method, FEL and SC are incorporated. The baseline method uses the same weights to fuse the feature maps generated by DAN, and the corresponding MAE is just 65.0. Specially, when FEL is introduced into the framework, the MAE is improved to 60.4. Thus, this approach is reasonable for fusing feature maps with learned weights. After SC is introduced into the framework, the MAE is improved to 59.2 justifying that SC is also an effective trick. The generated density maps are shown in Figure~\ref{fig9}. By comparing these density maps, we can further analyze the effects of different components in PaDNet. The density map generated by the baseline method is slightly blurred because it overestimates the count. Concretely, the bottom of the density map is biased. When FEL is employed to adjust the weights of feature maps, the generated density map becomes accurate and the overestimation at the bottom of the image is eliminated. However, the density map loses a little information in the middle of the image. When SC is employed, the lost information is supplemented.

\begin{table}[ht!]
\caption{The effects of different channel configurations of DAN on the UCF-QNRF dataset.}
 \renewcommand\arraystretch{1.2}
 \begin{tabular}{p{3.cm}<{\centering}|p{1.3cm}<{\centering}|p{1.3cm}<{\centering}}
 \hline

  Method & MAE & RMSE \\
 \hline
 \hline
 PaDNet-3-\textsl{ascend}& 106.7 & 184.0 \\

 PaDNet-3-\textsl{equal} & 103.5 & 179.7 \\

 PaDNet-3-\textsl{descend} & 101.8 & 180.8 \\

 PaDNet-3-\textsl{double} & 100.5 &  174.4 \\

 PaDNet-4 & \textbf{96.5} & \textbf{170.2} \\
 \hline

 \end{tabular}
 \centering
 \label{tab11}
 \end{table}
 \textbf{The effect of DAN.} We design DAN with  pyramidal filters to enhance the ability of capturing pan-density information. In particular, the lower-density subnetworks have relatively larger filters that is similar to  many multi-scale methods ~\cite{Zhang2016Single,sam2017switching,sindagi2017generating}. Hence, we mainly explore the effects of different channel configurations of DAN, and then conduct the studies on the UCF-QNRF dataset. The results are listed in Table~\ref{tab11}. We have three channel configurations for DAN of PaDNet-3, {\it i.e.}, \textsl{ascend}, \textsl{descend} and \textsl{equal}. The \textsl{descend} listed in Table~\ref{tab1} is our final design. The \textsl{ascend} means the lower subnetworks has fewer channels. The \textsl{equal} means all the subnetworks have the same number of channels as level-2 subnetwork (Table~\ref{tab1}).

Note that PaDNet-3-\textsl{descend} obtains the best recognition performance compared with PaDNet-3-\textsl{ascend} and PaDNet-3-\textsl{equal}. It confirms our speculation that lower-level subnetworks should equip with more filters than higher-level subnetworks in each layer because sparse scenes have more variations of crowds and environmental interference. When compared with sparse scenes, the distribution of dense crowds is closer to the uniform distribution. PaDNet-3-\textsl{ascend} has the fewest channels for level-1 subnetwork (Table~\ref{tab1}), hence it obtains the worst results.  Meanwhile, in order to exploit whether the number of parameters of DAN could greatly affect the final performance, we double the channels of level-3 subnetwork (Table~\ref{tab1}), and named as PaDNet-3-\textsl{double}, to make PadNet-3 and PaDNet-4 have same number of parameters. The MAE of PaDNet-3-\textsl{double} is 100.5 which is  better than PaDNet-3 because of the added parameters, but PaDNet-3-\textsl{double} still cannot achieve the competitive performance compared with PaDNet-4. Therefore, the number of subnetworks of DAN has a greater influence on final performance than does the number of parameters of DAN.
 \begin{figure*}[t]
 \centering
  \includegraphics[width=\linewidth, clip=true, trim=250 535 280 540]{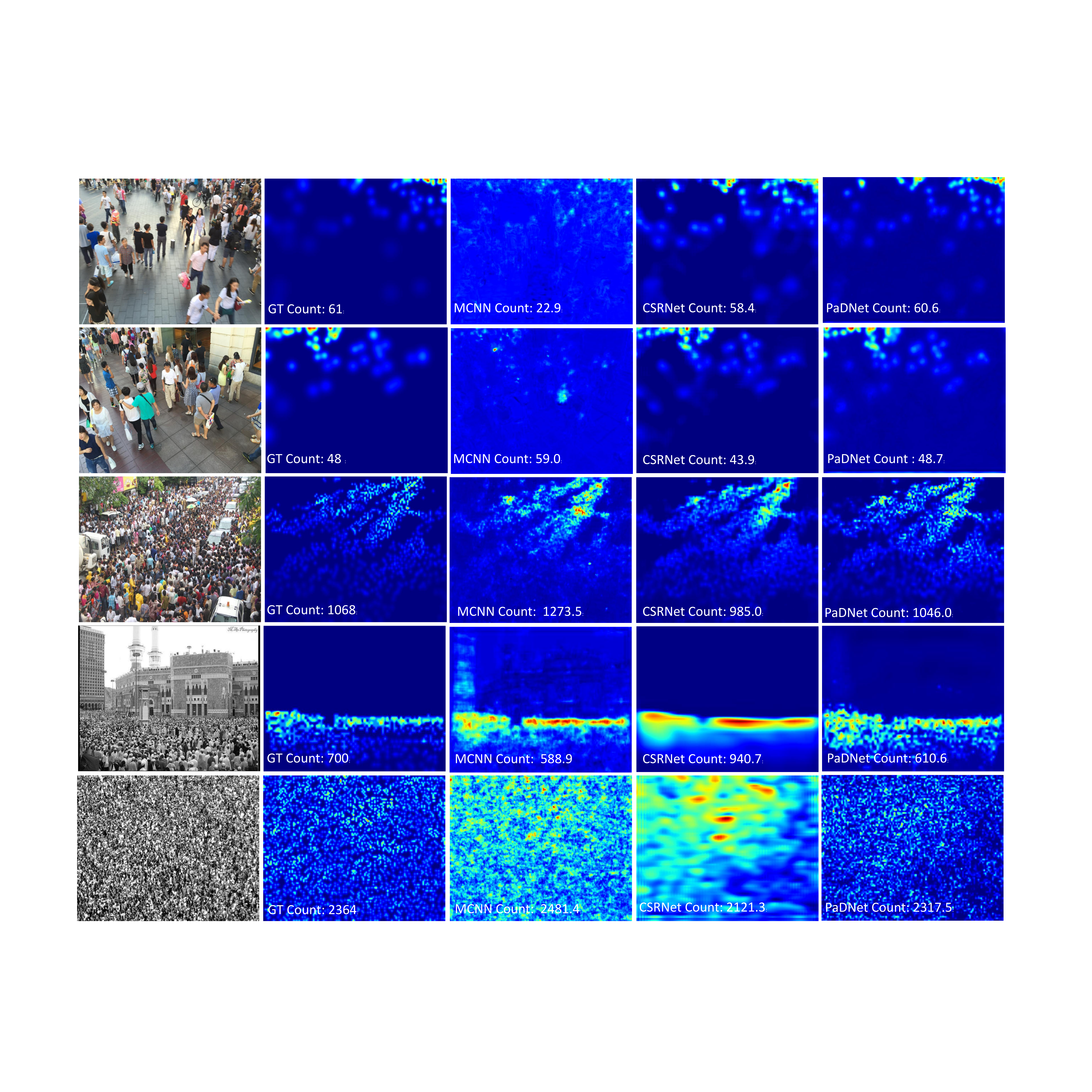}
 \caption{Example experimental results. The images in each row are original crowd image, the ground truth, the result generated by MCNN, the result generated by CSRNet\;(the state-of-the-art method based on single-column dilated convolutional network), and the result generated by our PaDNet, respectively. The images of the first two rows are in the SHB dataset. The images of the third row are in the SHA dataset. The remaining images are in the UCF\_CC\_50 dataset.}
 \label{fig11}
\end{figure*}
\begin{figure}[ht!]
 \centering
  \includegraphics[width=\linewidth, clip=true, trim=220 120 275 115]{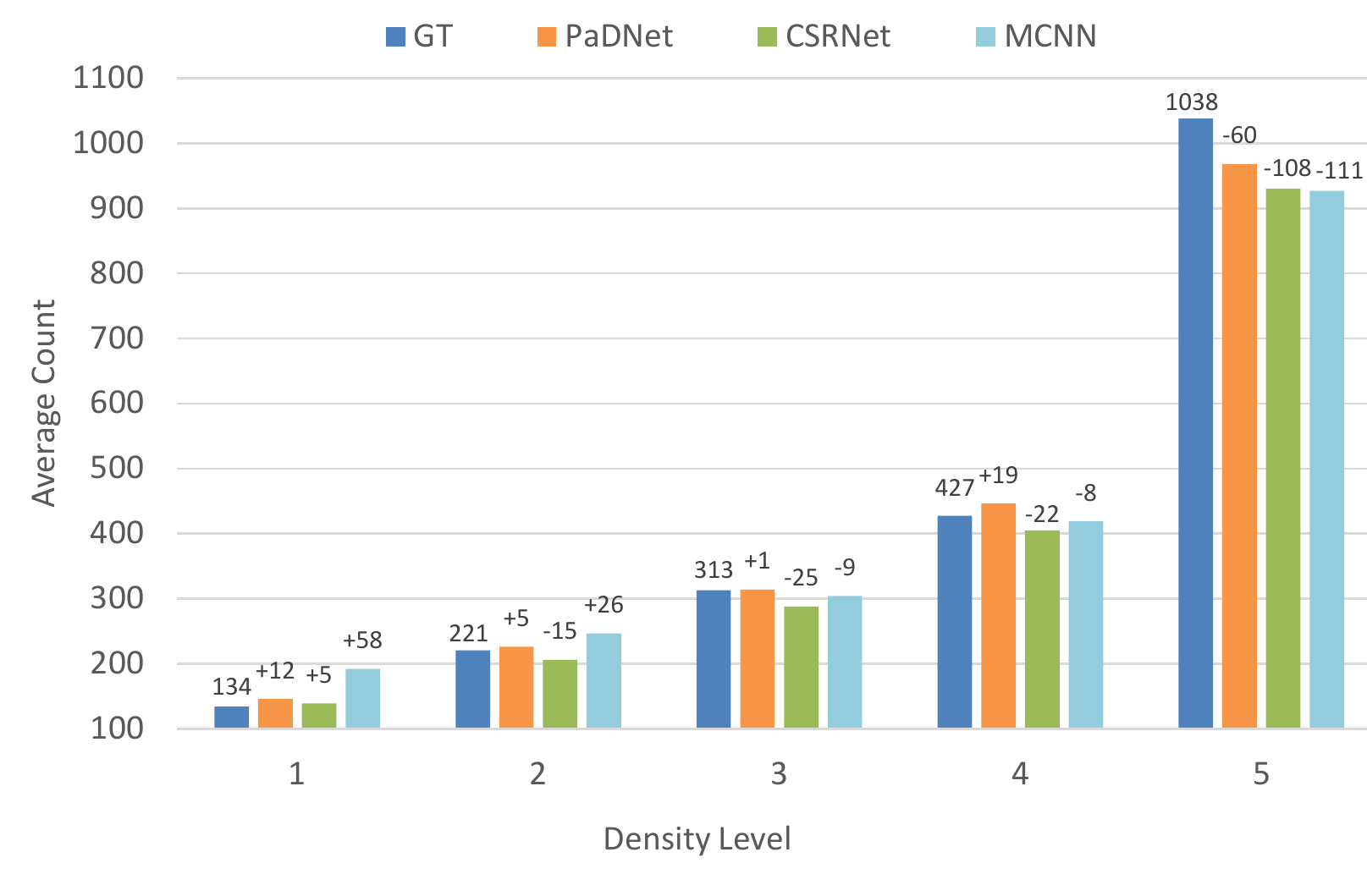}
 \caption{Histogram of average crowd numbers estimated by different methods on five groups splited from the SHA dataset according to increasing density level.}
 \label{fig10}
\end{figure}

  \textbf{The effect of weighting strategy for feature maps.}  A normal idea is that a network learns a $w_i$ to weight the $i$th feature map, but such a weighting strategy can only obtain limited performance improvement. Thus, we weight the importance for each feature map using Eq. \eqref{Equ3}. In order to verify the effectiveness of our strategy, we conduct the ablation studies on the SHA dataset with PaDNet-3. The results are listed in Table~\ref{tab12}. We discover that using $1 + w_i$ as the weight has a better performance than using $w_i$  as the weight. We argue that these feature maps generated by subnetworks are high-level features and quite close to the final density map.
   \begin{table}[ht!]
  \caption{The effects of different weight strategies for feature maps.}
 \renewcommand\arraystretch{1.2}
 \begin{tabular}{p{3.cm}<{\centering}|p{1.3cm}<{\centering}|p{1.3cm}<{\centering}}
 \hline

  Method & MAE & RMSE \\
 \hline
 \hline
 PaDNet($w_i$ as the weight)& 62.4 & 104.7 \\

 PaDNet($1 + w_i$ as the weight) & \textbf{59.2} & \textbf{98.1} \\

 \hline
 \end{tabular}
 \centering
 \label{tab12}
 \end{table}
 Here, $w_i$ is not the pixel-level weight and the value of $w_i$ ranging in $(0, 1)$. Using $w_i$  as the weight for the whole feature map could lose the local features of crowd. However, the form of $1 + w_{i}$ exhibits the combination of the original and the important features. Therefore, this strategy accurately weights the importance of feature maps at different levels based on $w_{i}$s and will not lose the original features.

\subsubsection{\textbf{Performance in pan-density crowd counting}}

We evaluate the performance of our method in pan-density crowd counting from two aspects: (i) the performance in different density scenes, and (ii) the performance at local regions of the same scene. We conduct the experiments on SHA with PaDNet-3. Meanwhile, we compare PaDNet with  MCNN and CSRNet\footnote{We implemented MCNN and CSRNet algorithms and obtained almost the same results.}.
In order to evaluate the performance in different density scenes, we divide the SHA dataset into five groups according to increasing density level. The results are shown in Figure~\ref{fig10}.
PaDNet achieves the best recognition performance on the  density levels 2, 3, and 5. CSRNet obtains the best performance on density level 1 and MCNN on level 4. However, PaDNet achieves competitive performance as CSRNet and MCNN on levels 1 and 4. Thus, PaDNet demonstrates both better accuracy and higher robustness in different density scenes.

As mentioned above, most existing methods performed well in global estimation while neglecting local accuracy. We evaluate the local accuracy and robustness for PaDNet according to proposed PMAE and PRMSE. We calculate PMAE and PRMSE when $n$ (Eq. \ref{PMAE} and \ref{PRMSE}.) is set to 1, 4, 9 and 16. The results are listed in Table~\ref{tab8}. The performance of PaDNet is better than MCNN and CSRNet under various conditions. {\it This suggests that regardless of global or local counts are examined, PaDNet achieves highly accurate and robust estimation in pan-density crowd counting.}

Meanwhile, we calculate PMAE and PRMSE of the ablated PaDNet. By comparing the last three rows of Table~\ref{tab8}, both FEL and SC are justified to be effective. Especially, for PaDNet without FEL and SC, the global MAE and RMSE are better than CSRNet. But the PMAE and PRMSE are worse than CSRNet. When FEL is introduced into the framework, the local evaluation is improved and the results are better than CSRNet. The experiments show that FEL is beneficial for improving the global and local recognition performance.

In order to evaluate the effectiveness of PMAE and PRMSE and to make it convenient for other researchers to follow pan-density crowd counting, we calculate proposed PMAE and PRMSE for MCNN, CSRNet, PaDNet-1, PaDNet-2, PaDNet-3, and PaDNet-4 on the SHA, SHB, UCF\_CC\_50, and UCF-QNRF datasets. The results are listed in Table~\ref{tab9}. It shows that PMAE and PRMSE are effective and robust metrics for evaluating the global and local accuracy and robustness. For example, in the SHA and UCF-QNRF datasets, the MAE of CSRNet is worse than PaDNet-2, while the PMAE of CSRNet is better than PaDNet-2. These results demonstrate that PMAE and PRMSE can effectively evaluate both global and local performances.

Figure~\ref{fig11} shows some density maps predicted by MCNN, CSRNet, and PaDNet. The density maps generated by MCNN are a little blurred, and the estimated counts are quite biased. Similarly, the density maps generated by CSRNet are also blurred in extremely dense scenes. In contrast, the density maps yielded by PaDNet indicate that not only is the local texture fine-grained but also the global one has high quality. Consequently, the counts of PaDNet are the closest to the ground truth.

Note that the trade-off for better performance is that data preprocessing is more complex because we must use different density datasets to pretrain the corresponding subnetworks. Furthermore, it takes about five hours to train the PaDNet on the ShanghaiTech Part\_A dataset using four NVIDIA GTX 1080Ti GPUs. In the prediction phase, the computation only costs 0.11 seconds on average for an image with one such GPU. Therefore, PaDNet can be readily deployed in real-time scene crowd counting.

\section{Conclusions}\label{sec:con}
We propose a novel end-to-end deep learning framework named PaDNet for pan-density crowd counting. PaDNet can fully leverage pan-density information. Specifically, the component DAN can effectively recognize different density crowds while the component FEL improves both global and local recognition performances. Meanwhile, the new evaluation metrics PMAE and PRMSE, which are extended from MAE and RMSE, not only evaluate the global accuracy and robustness but also the local ones. Extensive experiments on four benchmark datasets indicated that PaDNet attained the lowest predictive errors and higher robustness in pan-density crowd counting when compared with state-of-the-art algorithms. In the future, we will explore how to simplify network architecture for pan-density crowd counting.

\section*{Acknowledgments}
Y. Tian, Y. Lei, and J. Zhang were supported in part by the National Key R \& D Program of China (No. 2018YFB1305104), the National Natural Science Foundation of China (NSFC 61673118), Shanghai Municipal Science and Technology Major Project (No. 2018SHZDZX01) and ZJLab. J. Z. Wang was supported by The Pennsylvania State University. The authors would like to thank Hanqing Chao, Qi Zhou, Yuan Cao, and Haiping Zhu for their assistance. They are grateful to the reviewers and the Associate Editor for their constructive comments.


%
\bibliographystyle{IEEEtran}
\bibliography{references}

\begin{thebibliography}{10}
\providecommand{\url}[1]{#1}
\csname url@samestyle\endcsname
\providecommand{\newblock}{\relax}
\providecommand{\bibinfo}[2]{#2}
\providecommand{\BIBentrySTDinterwordspacing}{\spaceskip=0pt\relax}
\providecommand{\BIBentryALTinterwordstretchfactor}{4}
\providecommand{\BIBentryALTinterwordspacing}{\spaceskip=\fontdimen2\font plus
\BIBentryALTinterwordstretchfactor\fontdimen3\font minus
  \fontdimen4\font\relax}
\providecommand{\BIBforeignlanguage}[2]{{%
\expandafter\ifx\csname l@#1\endcsname\relax
\typeout{** WARNING: IEEEtran.bst: No hyphenation pattern has been}%
\typeout{** loaded for the language `#1'. Using the pattern for}%
\typeout{** the default language instead.}%
\else
\language=\csname l@#1\endcsname
\fi
#2}}
\providecommand{\BIBdecl}{\relax}
\BIBdecl

\bibitem{ford2017trump}
\BIBentryALTinterwordspacing
M.~Ford, ``Trump’s press secretary falsely claims: `largest audience ever to
  witness an inauguration, period','' \emph{The Atlantic}, January 21, 2017.
  [Online]. Available:
  \url{https://www.theatlantic.com/politics/archive/2017/01/\break
  inauguration-crowd-size/514058/}
\BIBentrySTDinterwordspacing

\bibitem{1541243}
B.~Wu and R.~Nevatia, ``Detection of multiple, partially occluded humans in a
  single image by {B}ayesian combination of edgelet part detectors,'' in
  \emph{Proceedings of the IEEE International Conference on Computer Vision},
  vol.~1, 2005, pp. 90--97.

\bibitem{5995698}
M.~Wang and X.~Wang, ``Automatic adaptation of a generic pedestrian detector to
  a specific traffic scene,'' in \emph{Proceedings of the IEEE Conference on
  Computer Vision and Pattern Recognition}, 2011, pp. 3401--3408.

\bibitem{chan2009bayesian}
A.~B. Chan and N.~Vasconcelos, ``Bayesian poisson regression for crowd
  counting,'' in \emph{Proceedings of the IEEE International Conference on
  Computer Vision}, 2009, pp. 545--551.

\bibitem{ryan2009crowd}
D.~Ryan, S.~Denman, C.~Fookes, and S.~Sridharan, ``Crowd counting using
  multiple local features,'' in \emph{Digital Image Computing: Techniques and
  Applications}, 2009, pp. 81--88.

\bibitem{8576646}
M.~{Kampffmeyer}, N.~{Dong}, X.~{Liang}, Y.~{Zhang}, and E.~P. {Xing},
  ``Connnet: A long-range relation-aware pixel-connectivity network for salient
  segmentation,'' \emph{IEEE Transactions on Image Processing}, vol.~28, no.~5,
  pp. 2518--2529, 2019.

\bibitem{8576656}
Y.~{Li}, J.~{Zeng}, S.~{Shan}, and X.~{Chen}, ``Occlusion aware facial
  expression recognition using cnn with attention mechanism,'' \emph{IEEE
  Transactions on Image Processing}, vol.~28, no.~5, pp. 2439--2450, 2019.

\bibitem{7839189}
K.~{Zhang}, W.~{Zuo}, Y.~{Chen}, D.~{Meng}, and L.~{Zhang}, ``Beyond a gaussian
  denoiser: Residual learning of deep cnn for image denoising,'' \emph{IEEE
  Transactions on Image Processing}, vol.~26, no.~7, pp. 3142--3155, 2017.

\bibitem{chao2019gaitset}
H.~Chao, Y.~He, J.~Zhang, and J.~Feng, ``{GaitSet}: Regarding gait as a set for
  cross-view gait recognition,'' in \emph{AAAI}, 2019.

\bibitem{Zhang2016Single}
Y.~Zhang, D.~Zhou, S.~Chen, S.~Gao, and Y.~Ma, ``Single-image crowd counting
  via multi-column convolutional neural network,'' in \emph{Proceedings of the
  IEEE Conference on Computer Vision and Pattern Recognition}, 2016, pp.
  589--597.

\bibitem{idrees2018composition}
H.~Idrees, M.~Tayyab, K.~Athrey, D.~Zhang, S.~Al-Maadeed, N.~Rajpoot, and
  M.~Shah, ``Composition loss for counting, density map estimation and
  localization in dense crowds,'' in \emph{Proceedings of the European
  Conference on Computer Vision}, 2018.

\bibitem{wang2015deep}
C.~Wang, H.~Zhang, L.~Yang, S.~Liu, and X.~Cao, ``Deep people counting in
  extremely dense crowds,'' in \emph{Proceedings of the ACM International
  Conference on Multimedia}, 2015, pp. 1299--1302.

\bibitem{Fu2015Fast}
M.~Fu, P.~Xu, X.~Li, Q.~Liu, M.~Ye, and C.~Zhu, ``Fast crowd density estimation
  with convolutional neural networks,'' \emph{Engineering Applications of
  Artificial Intelligence}, vol.~43, pp. 81--88, 2015.

\bibitem{8296324}
L.~Zeng, X.~Xu, B.~Cai, S.~Qiu, and T.~Zhang, ``Multi-scale convolutional
  neural networks for crowd counting,'' in \emph{Proceedings of the IEEE
  International Conference on Image Processing}, 2017, pp. 465--469.

\bibitem{sam2017switching}
D.~B. Sam, S.~Surya, and R.~V. Babu, ``Switching convolutional neural network
  for crowd counting,'' in \emph{Proceedings of the IEEE Conference on Computer
  Vision and Pattern Recognition}, 2017, pp. 4031--4039.

\bibitem{sindagi2017generating}
V.~Sindagi and V.~M.~Patel, ``Generating high-quality crowd density maps using
  contextual pyramid {CNN}s,'' in \emph{Proceedings of the IEEE International
  Conference on Computer Vision}, 2017, pp. 1879--1888.

\bibitem{li2018csrnet}
Y.~Li, X.~Zhang, and D.~Chen, ``{CSRN}et: Dilated convolutional neural networks
  for understanding the highly congested scenes,'' in \emph{Proceedings of the
  IEEE Conference on Computer Vision and Pattern Recognition}, 2018, pp.
  1091--1100.

\bibitem{Pedestrian}
P.~Dollar, C.~Wojek, B.~Schiele, and P.~Perona, ``Pedestrian detection: An
  evaluation of the state of the art,'' \emph{IEEE Transactions on Pattern
  Analysis and Machine Intelligence}, vol.~34, no.~4, pp. 743--761, 2012.

\bibitem{dalal2005histograms}
N.~Dalal and B.~Triggs, ``Histograms of oriented gradients for human
  detection,'' in \emph{Proceedings of the IEEE Conference on Computer Vision
  and Pattern Recognition}, vol.~1, 2005, pp. 886--893.

\bibitem{leibe2005pedestrian}
B.~Leibe, E.~Seemann, and B.~Schiele, ``Pedestrian detection in crowded
  scenes,'' in \emph{Proceedings of the IEEE Conference on Computer Vision and
  Pattern Recognition}, vol.~1, 2005, pp. 878--885.

\bibitem{tuzel2008pedestrian}
O.~Tuzel, F.~Porikli, and P.~Meer, ``Pedestrian detection via classification on
  {R}iemannian manifolds,'' \emph{IEEE Transactions on Pattern Analysis and
  Machine Intelligence}, vol.~30, no.~10, pp. 1713--1727, 2008.

\bibitem{felzenszwalb2010object}
P.~F. Felzenszwalb, R.~B. Girshick, D.~McAllester, and D.~Ramanan, ``Object
  detection with discriminatively trained part-based models,'' \emph{IEEE
  Transactions on Pattern Analysis and Machine Intelligence}, vol.~32, no.~9,
  pp. 1627--1645, 2010.

\bibitem{wu2007detection}
B.~Wu and R.~Nevatia, ``Detection and tracking of multiple, partially occluded
  humans by {B}ayesian combination of edgelet based part detectors,''
  \emph{International Journal of Computer Vision}, vol.~75, no.~2, pp.
  247--266, 2007.

\bibitem{Paragios2001}
N.~Paragios and V.~Ramesh, ``A {MRF}-based approach for real-time subway
  monitoring,'' in \emph{Proceedings of the IEEE Conference on Computer Vision
  and Pattern Recognition.}, vol.~1, 2001, pp. I--I.

\bibitem{Chan2008Privacy}
A.~B. Chan, Z.-S.~J. Liang, and N.~Vasconcelos, ``Privacy preserving crowd
  monitoring: Counting people without people models or tracking,'' in
  \emph{Proceedings of the IEEE Conference on Computer Vision and Pattern
  Recognition}, 2008, pp. 1--7.

\bibitem{Chen2012Feature}
K.~Chen, C.~L. Chen, S.~Gong, and T.~Xiang, ``Feature mining for localised
  crowd counting,'' in \emph{Proceedings of the British Machine Vision
  Conference}, 2012.

\bibitem{lempitsky2010learning}
V.~S. Lempitsky and A.~Zisserman, ``Learning to count objects in images,'' in
  \emph{Proceedings of the International Conference on Neural Information
  Processing Systems}, 2010, pp. 1324--1332.

\bibitem{Pham2015}
V.~Pham, T.~Kozakaya, O.~Yamaguchi, and R.~Okada, ``Count forest: Co-voting
  uncertain number of targets using random forest for crowd density
  estimation,'' in \emph{Proceedings of the IEEE International Conference on
  Computer Vision}, 2015, pp. 3253--3261.

\bibitem{zhang2015cross}
C.~Zhang, H.~Li, X.~Wang, and X.~Yang, ``Cross-scene crowd counting via deep
  convolutional neural networks,'' in \emph{Proceedings of the IEEE Conference
  on Computer Vision and Pattern Recognition}, 2015, pp. 833--841.

\bibitem{Sindagi2017}
V.~A. Sindagi and V.~M. Patel, ``{CNN}-based cascaded multi-task learning of
  high-level prior and density estimation for crowd counting,'' in
  \emph{Proceedings of the IEEE International Conference on Advanced Video and
  Signal Based Surveillance}, 2017, pp. 1--6.

\bibitem{kang2018crowd}
D.~Kang and A.~Chan, ``Crowd counting by adaptively fusing predictions from an
  image pyramid,'' in \emph{Proceedings of the British Machine Vision
  Conference}, 2018.

\bibitem{deb2018aggregated}
D.~Deb and J.~Ventura, ``An aggregated multicolumn dilated convolution network
  for perspective-free counting,'' in \emph{Proceedings of the IEEE Conference
  on Computer Vision and Pattern Recognition Workshops}, 2018, pp. 195--204.

\bibitem{Cao_2018_ECCV}
X.~Cao, Z.~Wang, Y.~Zhao, and F.~Su, ``Scale aggregation network for accurate
  and efficient crowd counting,'' in \emph{Proceedings of the The European
  Conference on Computer Vision}, 2018.

\bibitem{Shi2015Convolutional}
X.~Shi, Z.~Chen, H.~Wang, D.~Y. Yeung, W.~Wong, and W.~Woo, ``Convolutional
  {LSTM} network: A machine learning approach for precipitation nowcasting,''
  in \emph{Proceedings of the International Conference on Neural Information
  Processing Systems}, 2015, pp. 802--810.

\bibitem{Xiong2017Spatiotemporal}
F.~Xiong, X.~Shi, and D.~Yeung, ``Spatiotemporal modeling for crowd counting in
  videos,'' in \emph{Proceedings of the IEEE International Conference on
  Computer Vision}, 2017, pp. 5161--5169.

\bibitem{shen2018crowd}
Z.~Shen, Y.~Xu, B.~Ni, M.~Wang, J.~Hu, and X.~Yang, ``Crowd counting via
  adversarial cross-scale consistency pursuit,'' in \emph{Proceedings of the
  IEEE Conference on Computer Vision and Pattern Recognition}, 2018, pp.
  5245--5254.

\bibitem{liu2018decidenet}
J.~Liu, C.~Gao, D.~Meng, and A.~G. Hauptmann, ``{D}ecide{N}et: Counting varying
  density crowds through attention guided detection and density estimation,''
  in \emph{Proceedings of the IEEE Conference on Computer Vision and Pattern
  Recognition}, 2018, pp. 5197--5206.

\bibitem{shi2018crowd}
Z.~Shi, L.~Zhang, Y.~Liu, X.~Cao, Y.~Ye, M.-M. Cheng, and G.~Zheng, ``Crowd
  counting with deep negative correlation learning,'' in \emph{Proceedings of
  the IEEE Conference on Computer Vision and Pattern Recognition}, 2018, pp.
  5382--5390.

\bibitem{Liu_2018_CVPR}
X.~Liu, J.~van~de Weijer, and A.~D. Bagdanov, ``Leveraging unlabeled data for
  crowd counting by learning to rank,'' in \emph{Proceedings of the IEEE
  Conference on Computer Vision and Pattern Recognition}, 2018.

\bibitem{Ranjan_2018_ECCV}
V.~Ranjan, H.~Le, and M.~Hoai, ``Iterative crowd counting,'' in
  \emph{Proceedings of the The European Conference on Computer Vision}, 2018.

\bibitem{vgg}
\BIBentryALTinterwordspacing
K.~Simonyan and A.~Zisserman, ``Very deep convolutional networks for
  large-scale image recognition,'' \emph{CoRR}, vol. abs/1409.1556, 2014.
  [Online]. Available: \url{http://arxiv.org/abs/1409.1556}
\BIBentrySTDinterwordspacing

\bibitem{resnet}
K.~He, X.~Zhang, S.~Ren, and J.~Sun, ``Deep residual learning for image
  recognition,'' in \emph{IEEE Conference on Computer Vision and Pattern
  Recognition}, 2016, pp. 770--778.

\bibitem{googlenet}
C.~Szegedy, W.~Liu, Y.~Jia, P.~Sermanet, S.~Reed, D.~Anguelov, D.~Erhan,
  V.~Vanhoucke, and A.~Rabinovich, ``Going deeper with convolutions,'' in
  \emph{Proceedings of the IEEE Conference on Computer Vision and Pattern
  Recognition}, 2015, pp. 1--9.

\bibitem{Russakovsky2015ImageNet}
O.~Russakovsky, J.~Deng, H.~Su, J.~Krause, S.~Satheesh, S.~Ma, Z.~Huang,
  A.~Karpathy, A.~Khosla, and M.~Bernstein, ``Image{N}et large scale visual
  recognition challenge,'' \emph{International Journal of Computer Vision},
  vol. 115, no.~3, pp. 211--252, 2015.

\bibitem{NIPS2014_5347}
J.~Yosinski, J.~Clune, Y.~Bengio, and H.~Lipson, ``How transferable are
  features in deep neural networks?'' in \emph{Advances in Neural Information
  Processing Systems 27}, 2014, pp. 3320--3328.

\bibitem{BN}
S.~Ioffe and C.~Szegedy, ``Batch {N}ormalization: Accelerating deep network
  training by reducing internal covariate shift,'' \emph{CoRR}, vol.
  abs/1502.03167, 2015.

\bibitem{he2014spatial}
K.~He, X.~Zhang, S.~Ren, and J.~Sun, ``Spatial pyramid pooling in deep
  convolutional networks for visual recognition,'' \emph{IEEE Transactions on
  Pattern Analysis and Machine Intelligence}, vol.~37, no.~9, pp. 1904--1916,
  2015.

\bibitem{unet}
O.~Ronneberger, P.~Fischer, and T.~Brox, ``U-{N}et: Convolutional networks for
  biomedical image segmentation,'' in \emph{Proceedings of the International
  Conference on Medical Image Computing and Computer-Assisted Intervention},
  2015, pp. 234--241.

\bibitem{densehuang2017}
G.~Huang, Z.~Liu, L.~v.~d. Maaten, and K.~Q. Weinberger, ``Densely connected
  convolutional networks,'' in \emph{Proceedings of the IEEE Conference on
  Computer Vision and Pattern Recognition}, 2017, pp. 2261--2269.

\bibitem{Idrees2013Multi}
H.~Idrees, I.~Saleemi, C.~Seibert, and M.~Shah, ``Multi-source multi-scale
  counting in extremely dense crowd images,'' in \emph{Proceedings of the IEEE
  Conference on Computer Vision and Pattern Recognition}, 2013, pp. 2547--2554.

\bibitem{Guerrero2015Extremely}
R.~Guerrero-Gómez-Olmedo, B.~Torre-Jiménez, R.~López-Sastre,
  S.~Maldonado-Bascón, and D.~Oñoro-Rubio, ``Extremely overlapping vehicle
  counting,'' in \emph{Iberian Conference on Pattern Recognition and Image
  Analysis}, 2015.

\bibitem{Sam_2018_CVPR}
D.~Babu~Sam, N.~N. Sajjan, R.~Venkatesh~Babu, and M.~Srinivasan, ``Divide and
  grow: Capturing huge diversity in crowd images with incrementally growing
  {CNN},'' in \emph{Proceedings of the IEEE Conference on Computer Vision and
  Pattern Recognition}, 2018.

\bibitem{huangbody}
S.~Huang, X.~Li, Z.~Zhang, F.~Wu, S.~Gao, R.~Ji, and J.~Han, ``Body structure
  aware deep crowd counting,'' \emph{IEEE Transactions on Image Processing},
  vol.~27, no.~3, pp. 1049--1059, 2018.

\bibitem{segnet}
V.~Badrinarayanan, A.~Kendall, and R.~Cipolla, ``Seg{N}et: A deep convolutional
  encoder-decoder architecture for image segmentation,'' \emph{IEEE
  Transactions on Pattern Analysis and Machine Intelligence}, vol.~39, no.~12,
  pp. 2481--2495, 2017.

\end{thebibliography}

\newpage
\end{document}